\documentclass[a4paper,12pt]{article}
\usepackage[left=1.3cm,right=1.3cm,top=2.5cm,bottom=2.5cm]{geometry}
\usepackage{setspace}
\usepackage{amsthm}
\usepackage{amssymb}
\usepackage{amsmath}
\usepackage{algorithm}
\usepackage{algorithmic}
\usepackage{tikz}
\usetikzlibrary{arrows}
\usetikzlibrary{shapes}

\usepackage{natbib}
 \bibpunct[, ]{(}{)}{,}{a}{}{,}%
 \def\newblock{\ }%
\newcommand{\card}[1]{\left\vert{#1}\right\vert}
\newcommand{\nop}[1]{}

\def\argmin{\mathop{\rm arg\,min}}%

\newtheorem{theorem}{Theorem}
\newtheorem{lemma}{Lemma}
\newtheorem{example}{Example}
\newtheorem{definition}{Definition}

\begin{document}
\onehalfspacing

\title{Fair task allocation in transportation}
\author{
Qing Chuan Ye\textsuperscript{a}, Yingqian Zhang\textsuperscript{b}, Rommert Dekker\textsuperscript{a} \\
\small \textsuperscript{a}Erasmus University Rotterdam, P.O. Box 1738, 3000 DR, Rotterdam, The Netherlands \\
\small \textsuperscript{b}Eindhoven University of Technology, P.O. Box 513, 5600 MB, Eindhoven, The Netherlands\\
}
\date{}

\maketitle

\abstract{
Task allocation problems have traditionally focused on cost optimization. However, more and more attention is being given to cases in which cost should not always be the sole or major consideration.
In this paper we study a fair task allocation problem in transportation where an optimal allocation not only has low cost but more importantly,
it distributes tasks as even as possible among heterogeneous participants who have different capacities and costs to execute tasks.
To tackle this fair minimum cost allocation problem we analyze and solve it in two parts using two novel polynomial-time algorithms.
We show that despite the new fairness criterion, the proposed algorithms can solve the fair minimum cost allocation problem optimally in polynomial time.
In addition, we conduct an extensive set of experiments to investigate the trade-off between cost minimization and fairness. Our experimental results demonstrate the benefit of factoring fairness into task allocation. Among the majority of test instances, fairness comes with a very small price in terms of cost.  
}

\section{Introduction}\label{intro}
Traditionally, optimization of task allocation problems considered only the costs involved in the allocation. However, there has been in recent years more attention to cases where cost should not always be the sole consideration \citep{campbell2008routing}. There are circumstances when other criteria need to be taken into account as well during the decision making process. \emph{Fairness} has been considered as one of the important additional criteria in many application domains \citep{ogryczak2005telecommunications, gopinathan2011strategyproof, bertsimas2012efficiency}.
Although there is no common definition for the term, there are two fairness criteria that are often used in the literature: the Nash bargaining criterion and the Rawlsian maximin criterion. The former is based on Nash's four axioms of pareto optimality, independency of irrelevant alternatives, symmetry, and invariance to affine transformations or equivalent utility representations \citep{nash1950bargaining}. The latter is based on Rawls' two principles of justice \citep{rawls2009theory}. Rawls' maximin criterion maximizes the welfare level of the worst-off group member and has therefore been used in allocation problems \citep{jaffe1981bottleneck,kumar2000fairness}.


In this paper, we study task allocation problems in which we take fairness into account in addition to the standard minimum cost criterion. This work was inspired by an actual transportation situation in the port of Rotterdam in the Netherlands. The increase in the number of container terminals in the said port will result in a huge increase in inter-terminal transport (ITT). The port authority invited a team of researchers to investigate a sustainable transportation system, called an \emph{asset light solution}, in which trucks that were already present in the port could execute open jobs. The main idea behind this system is that trucks that come from the hinterland to drop off or pick up containers often have spare time in between tasks. Usually, trucks are scheduled to do several jobs to and from various terminals in the port in one day. There may be large gaps between these jobs during which time the truck would be idle due to the nature of the jobs that truck companies agree to do. Terminals could take advantage of these idle trucks by providing them with jobs that they can perform within the port while waiting for their next scheduled job. The trucks will be compensated for these jobs. The compensation from the terminals to the trucking companies would be large enough to cover the costs that the companies would incur. However, the compensation should be less than the costs of purchasing and maintaining, or even renting the vehicles dedicated for such jobs. This way, the trucking companies gain additional income while the terminals save money by using readily available resources. Furthermore, because the utilization rate of existing trucks becomes higher and no new trucks are needed, this is a more durable approach to meeting the transport need within the port.

To realize such a task allocation, terminals need to be informed of the individual schedules of the different trucking companies. This poses a hurdle because getting such information is expensive and the trucking companies may be reluctant to share their entire schedules. 
One way to circumvent this difficulty is to use \textit{auctions} as a means to collect information from different parties.
Auctions have become increasingly popular for allocating resources among individual players in many application domains, such as in spectrum auctions \citep{Cramton02}, health care \citep{smits2008impact}, industrial procurement \citep{gallien2005smart, bichler2006industrial} and logistics \citep{sheffi2004combinatorial, ball2006auctions}.
In the auction for our trucking task allocation case, we assume that all terminals together act as an auctioneer and they announce a set of available jobs. Different trucking companies can bid for those jobs, depending on their idle trucks at specific times.
Given the bids of different companies, the terminals then decide on a best allocation of jobs to companies.\footnote{Auctions are used in this research as a way to collect local information from the participants. 
We do not consider the bidding behaviour of the bidders in this paper.}     
Because there are ITT movements every day that need to be executed, this task allocation activity would be held daily. Some studies have shown that greedily minimizing cost does not fare well with repeated auctions. Participants could experience starvation in the long run, which will reduce their incentive to continue participating in the allocation activity \citep{gopinathan2011strategyproof}. 
Furthermore, repeated auctions may affect the relationships between the auctioneer and bidders, which in turn affects the latter's way of bidding \citep{jap2008interorganizational}. To prevent these adverse effects, we should not only look at optimizing the costs in the task allocation, but we should also incorporate fairness in the task allocation that results from the auctions. We do this by reassuring that all interested parties will receive some market share, therefore giving trucking companies an incentive to continue participating in the task allocation activity. As we do not know the exact utility functions of the players, the number of jobs allocated to them will be used to measure the fair distribution of the utilities of the players.

We study a ``max-min fair minimum cost allocation  problem'' (MFMCA).
The majority of existing work involving fairness uses mathematical programming models in which fairness is incorporated in either the constraints~\citep{meng2002benefit, perugia2011designing} or in the objective function~\citep{bertsimas2011integer, bertsimas2012efficiency, barnhart2012equitable}. However, we aim for a polynomial-time solution.
The difficulty of our problem lies in the additional fairness criterion, which requires the developed algorithm to satisfy three criteria: allocation maximization, fairness, and cost minimization. To the best of our knowledge, no existing polynomial-time algorithm can be directly applied to solve our problem.
In this paper, we propose polynomial-time algorithms to solve MFMCA as a two-level optimization problem. First, we aim at a fairest allocation among companies while ensuring that a maximal set of tasks can be allocated for execution. We call this the ``max-min fair allocation problem'' (MMFA). Second, because there might be an exponential number of allocations that are considered max-min fair, we would like to determine which of these fair allocations has the lowest cost. The resulting allocation is max-min fair with minimum cost.
To this end, we develop a polynomial-time optimal method that consists of two novel algorithms: (1) to solve MMFA, we construct an algorithm, called IMaxFlow, using a progressive filling idea in a flow network \citep{bertsekas1987data}, and then (2) by using the solution obtained from MMFA, we propose another algorithm, called FairMinCost, that smartly alters the structure of the problem to solve MFMCA optimally.


The contribution of this paper is two-fold.
\begin{itemize}
\item[1.] Despite the new fairness criterion, we are able to develop an optimization method  to solve the task allocation problem to optimality in polynomial time.
\item[2.] Using computational results, we provide insights into situations in which fairness can be incorporated without giving up too much efficiency.
\end{itemize}

The rest of the paper is organized as follows.
We start with a literature review in Section~\ref{sec:review}, followed by a problem definition in Section~\ref{sec:def}. In Section~\ref{sec:opt}, we introduce two polynomial-time algorithms to solve MMFA and MFMCA, respectively. We prove that the output of these algorithms is the optimal allocation in terms of fairness and cost minimization. In Section~\ref{sec:computational}, using different sets of scenarios, we test the algorithm in terms of its effect on the cost and job distribution. We conclude and point out interesting directions of future work in Section~\ref{sec:con}.

\section{Literature review}
\label{sec:review}

The idea of factoring fairness into decision making has been studied in various fields. One of the earlier and still important areas of application where fairness has been considered is that of bandwidth allocation in telecommunication networks \citep{jaffe1981bottleneck, zukerman2005efficiency}. In this area, continuous flows with predefined origin-destination pairs are used, leading to algorithms that increase flow over all paths simultaneously until links are saturated, or that split up bandwidth equally among competitors. \cite{bertsekas1987data} give a simple algorithm for computing max-min fair rate vectors for flow control in networks, the so-called progressive filling algorithm, which is treated as one of the standard fairness concepts within the telecommunications or network applications \citep{ogryczak2005telecommunications}. In their problem setting, they assume that each session has an associated fixed path in the network. The algorithm starts with no flow, and then flow gets gradually increased over all paths simultaneously until a link in a path is saturated. The algorithm then continues from step to step, equally incrementing the flow in all paths that are not using saturated links, until all paths contain at least one saturated link.
\cite{tomaszewski2005polynomial} provides a general mathematical programming formulation for solving max-min fair problems using the progressive filling algorithm. 
Although we cannot use these proposed solution methods directly, we are able to borrow the idea of the progressive filling algorithm when developing our method for solving MMFA.

Fairness, or equity, has also been incorporated in staff scheduling. They attempt to distribute the workload fairly and evenly among personnel, where it is a typical strategy to construct cyclic rosters \citep{ernst2004staff}. The more popular measures for equity in this field are the variance and variants of the Gini index. Equity is then incorporated in mathematical models in either the objective function, e.g. minimizing the variation in workload, or through the use of constraints, which provide lower and upper bounds on the workload \citep{eiselt2008employee}.
Resource allocation is yet another field in which fairness plays an important role. An example of a very weak fairness constraint in this field is that any task will be able to use its requested resource eventually. A much stricter fairness requirement can be found in proportionate fairness \citep{baruah1996proportionate}. With proportionate fairness, the difference in the number of resource allocations to tasks will never be more than one, ensuring that all tasks have similar access to resources. Dominant Resource Fairness is another type of fairness requirement, which is a generalization of max-min fairness for multiple resources, where it maximizes the minimum dominant share across all users \citep{ghodsi2011dominant}. Fairness influences the order in which resources are scheduled to tasks, as certain tasks may take precedence.

Another domain in which fairness is incorporated is the field of air traffic management. In this field, fairness is important for air traffic flow management \citep{lulli2007european,barnhart2012equitable}, flight scheduling \citep{kubiak2009proportional}, and allocation of take-off and landing slots at airports \citep{bertsimas2011integer,bertsimas2012efficiency}. These studies consider a fair distribution of the utilities of all players usually expressed in monetary units or delay time. The air traffic flow management problem has been shown to be NP-hard \citep{bertsimas1998air}, and therefore mathematical programming models are often used in which the fairness measurement is incorporated in the objective function with which good computational results are achieved \citep{bertsimas2011integer, bertsimas2012efficiency, barnhart2012equitable}.
In addition, \cite{hoffman2005resource} and \cite{kim2015some} emphasize that equity and fairness are important in the air traffic flow management program design, because equitable treatment of airlines in such programs will be less likely to encourage gaming behaviour by a highly competitive industry. If one fails to consider equity, it might be detrimental to an otherwise well-designed air traffic flow management program.
\cite{ogryczak2014fair} provides a nice overview of the various areas of application of fairness and the most important models and methods of fair optimization.

Surprisingly, fairness has not yet been investigated widely in transportation optimization problems, although it has been treated as a psychological factor that influences acceptability of policies like road pricing \citep{fujii2004cross,eriksson2008acceptability}.
In road network design fairness is also an issue, because without fairness network users might not get any benefit from the network design project, and therefore it may be difficult to rally public support and it may be easy to evoke opposition to the implementation of such a project \citep{meng2002benefit}. In this application fairness is enforced through the addition of a constraint on the difference of the travel cost ratio between before and after the project.
There has also been some work in vehicle routing problems, where fairness is considered in the extra-time distribution of a transportation service \citep{perugia2011designing}. In order to incorporate fairness, they make use of a capping function, which enforces an upper bound on the extra-time.
\cite{litman2002evaluating} gives an overview of many different transportation decisions where fairness could be incorporated.
However, there is hardly any literature on incorporating fairness in task allocation problems in transportation.




We use the number of tasks allocated to a player as our measure of fairness. Thus, fairness is a property inherent in the allocation itself. We introduce a novel solution method because in our problem we try to assign tasks to players without any information on the players' utilities. This is in contrast to \cite{bertsimas2011price, bertsimas2012efficiency}, who assume that one knows the utilities of players, such that efficiency and fairness can be expressed as a function of the utilities.
In addition, we define our fairness measurement in terms of the allocation itself rather than in terms of some characteristic of the consequence of the allocation. Examples of the latter are tardiness and delay time, which are often used in air traffic flow management~\citep{lulli2007european, bertsimas2011integer, barnhart2012equitable}.
We will show that our fairness measurement simplifies the optimization problem and that we are able to develop polynomial-time algorithms to find an optimal fair allocation, which is highly desirable in practice.

\section{Problem definition}
\label{sec:def}

We assume that the set of available tasks (or jobs) to be distributed is known in advance by the central planner. For instance, in our motivating example, the terminals know a day in advance which container vessels will arrive and how many containers they will need to handle. The terminals are thus able to construct a schedule for their vehicles and cranes a day ahead, and this schedule reveals the necessary inter-terminal transport movements. These inter terminal transport movements are the jobs to be auctioned.
We assume a set of time periods $\mathcal{T}$, which consists of $T$ time periods. The set of jobs, denoted by $\mathcal{J}$ consisting of a total of $J$ jobs, comes with an earliest available time and a latest completion time for each job.
We assume that jobs are independent.\footnote{The relaxation of this independence assumption will be discussed in Section~\ref{sec:con}.} Each job can therefore be executed individually regardless of the execution of other jobs.
We define for each job $j_i\in \mathcal{J}$ its possible starting time as a mapping: $\mathcal{J}\times \mathcal{T} \mapsto \{0,1\}$. When it is clear from the context, we abuse the notation and use $j_i^t$ to denote that job $j_i$ is available at time period $t \in \mathcal{T}$.

Once the set of jobs $\mathcal{J}$ together with their possible starting times has been made available, a set of companies $\mathcal{K}$, consisting of $K$ companies, may bid on individual jobs. We do not consider combinatorial bids in this paper.
In addition to the selection of jobs that a company $k\in \mathcal{K}$ wishes to perform, the company also needs to provide 
their available capacity $n_k^t$ in time period $t$ in which it is able to perform the jobs.
We assume that each job takes up one unit of capacity and can be completed within one time unit. Furthermore, the company $k$ needs to provide its desired compensation (or cost), $c(j_i,k)$, for the bid job $j_i \in \mathcal{J}$.  A bid, $B_k$, from a company $k$ is thus a tuple: $\langle \boldsymbol{c}_k, \boldsymbol{n}_k \rangle$, where $\boldsymbol{c}_k$ is a set of costs $c(j^t_i,k)$, which denote the compensation of performing job $j_i$ at time $t$, and $\boldsymbol{n}_k$ is a set of capacities $n^t_k$, which specify the capacity of company $k$ at time period $t$.

The focus of this research is on the design of  task allocation algorithms, and not on the auction design. Therefore, to illustrate our approach, we adopt a simple
sealed-bid first-price auction format, where terminals can announce their available tasks and each company could submit their bids via, for example, a bidding website. In a sealed-bid first-price auction all bidders submit their sealed bids simultaneously so that no bidder knows the bids of other participants and the winning bidder pays the price they submitted. Once all bids from the bidding companies $\mathcal{K}$ have been collected, which can be enforced by a time limit, the auctioneer then decides which companies get to execute which jobs, that is, the auctioneer determines a task allocation
 $\pi: \mathcal{J}\times \mathcal{T} \times \mathcal{K} \mapsto \{0,1\}$.  An allocation is \emph{feasible} if (1) each job is allocated to at most one time slot and to at most one company, i.e., for each $j\in \mathcal{J}$, $\sum_{t\in \mathcal{T}}\sum_{k\in \mathcal{K}}\pi(j,t,k) \leq 1$; and (2) the number of jobs needed to be executed at time $t$ does not exceed the capacity at time $t$ of the company to which those jobs are assigned, i.e., for each $k$ and $t$, $\sum_{j \in \mathcal{J}} \pi(j,t,k) \leq n^t_k$.
The companies then receive the corresponding compensations specified in their bids for executing the assigned jobs.
The focus of this paper is on determining an optimal allocation $\pi$ of jobs to bidders. Following our motivating example, there are three ordered objectives for a fair job allocation: (1) the number of allocated jobs in $\pi$ is maximized,  (2) the allocation is fair to the bidders, and (3) the total compensation for executing the jobs is minimized.

Objectives 1 and 3 are rather straightforward given the context. For the fairness objective,
we use the notion of max-min fairness derived from Rawls' fairness principle \citep{rawls2009theory}. The central idea of max-min fairness is that the minimum utility of all bidders will have been maximized.
In this work,
we use the number of allocated jobs as a measure of the bidders' utility. In this way, we do not need to worry about the companies' actual utility functions, which they are likely unwilling to share with the auctioneer and which are difficult to model.

Given a feasible allocation $\pi$, let the number of allocated jobs be $Z=\card{\{j_i\ : j_i\in \mathcal{J}, \pi(j_i,\cdot,\cdot)=1\}}$. 
Let $\boldsymbol{\omega} = (\omega_1, \dots, \omega_{K})$ denote the number of jobs $\omega_k$ assigned to company $k \in \mathcal{K}$ in $\pi$. We call $\boldsymbol{\omega}$ an allocation vector. Clearly, it holds that $\sum_{\omega_k\in \boldsymbol{\omega}} \omega_k = Z$. Given $Z$ jobs, there may exist many possible allocations that distribute $Z$ jobs to $K$ companies. We call an allocation vector $\boldsymbol{\omega}$ \emph{$Z$-feasible} if and only if $\boldsymbol{\omega}$ can lead to a feasible allocation and $\sum_{\omega_k\in \boldsymbol{\omega}} \omega_k = Z$.

The max-min fairness principle entails that given a total of $Z$ jobs, the number of jobs for any company cannot be increased by at the same time decreasing the number of jobs of the other companies that have the same number of jobs or less. More formally, let $\boldsymbol{\omega}$ be a $Z$-feasible vector, and $\sigma$ be a sorting operator in which the components of $\boldsymbol{\omega}$ are sorted in nondecreasing order: $\sigma(\boldsymbol{\omega})_i \leq \sigma(\boldsymbol{\omega})_j$ if $\omega_i \leq \omega_j$. Let $\boldsymbol{\phi} = \sigma(\boldsymbol{\omega})$. We want to maximize the lexicographical minimum in all $Z$-feasible allocation vectors $\boldsymbol{\phi}$.
Intuitively speaking, we want to have an allocation that distributes a set of jobs among the companies as evenly as possible.

\begin{definition}[Max-min fairness]\label{def:fair}
Given $Z$ jobs to be distributed, we say a $Z$-feasible sorted allocation vector $\boldsymbol{\phi}$ is \emph{lexicographically greater} than another $Z$-feasible sorted vector $\boldsymbol{\phi}'$
if there exists a smallest index $j$ ($1 \leq j \leq K$) such that $\boldsymbol{\phi}_j > \boldsymbol{\phi}'_j$, and for index $i$, $1\leq i<j$, it holds that $\boldsymbol{\phi}_i = \boldsymbol{\phi}'_i$.
An allocation vector is \emph{max-min fair} with regard to $Z$ jobs if it is lexicographically greater than any other $Z$-feasible vector.
\end{definition}

We now use the following example to illustrate the three objectives of the job allocation problem.

\begin{example}\label{example_MLFA}

Suppose we have $5$ jobs to be auctioned. The jobs can be done in the following time periods: $(j_1: j_1^1);(j_2: j_2^2, j_2^4); (j_3: j_3^2, j_3^3); (j_4: j_4^3, j_4^4); (j_5: j_5^5)$. Three companies submit their bids, as shown in Table~\ref{tab:MLFA}.
The first row in the table shows that company $k_1$ bids on job $j_1$ that is to be executed during time period 1, for a compensation of $20$.

\begin{table}
\centering
{\scriptsize
\begin{tabular}{|l|l|l|l|l|l|}
\hline
Time points 	& 1 & 2 & 3 & 4 & 5 \\
\hline
company $k_1$  	&  $j_1: 20$  &  &  &  &   \\
company $k_2$  	&  $j_1: 30$  & $j_2: 40$, $j_3: 25$ &  &   &    \\
company $k_3$  	&  $j_1: 10$  & $j_2: 20$, $j_3: 20$   & $j_3: 25$, $j_4: 25$ & $j_2: 30$, $j_4: 20$  & $j_5: 20$  \\
\hline
\end{tabular}
}
\caption{The bids of three companies include desired jobs in each time period and their associated costs. The capacity of all companies is assumed to be 1 for each time period.
\label{tab:MLFA}}
{}
\end{table}

In this example, all $5$ jobs can be feasibly assigned. There are five feasible allocations: $\pi_1$ assigns
$j_1^1, j_2^2$ to $k_2$ and $j_3^3, j_4^4, j_5^5$ to $k_3$;
 $\pi_2$   assigns $j_1^1$ to $k_1$ and $j_2^2, j_3^3, j_4^4, j_5^5$ to $k_3$;
 $\pi_3$ assigns $j_3^2$ to $k_2$ and $j_1^1, j_2^2, j_4^4, j_5^5$ to $k_3$;
 $\pi_4$ assigns $j_1^1$ to $k_1$, $j_2^2$ to $k_2$ and $j_3^3, j_4^4, j_5^5$ to $k_3$; and $\pi_5$ assigns $j_1^1$ to $k_1$, $j_3^2$ to $k_2$ and $j_2^2, j_4^4, j_5^5$ to $k_3$.
The allocation vectors of these five assignments are $\boldsymbol{\phi}_1=(0,2,3)$, $\boldsymbol{\phi}_2=\boldsymbol{\phi}_3=(0,1,4)$, and $\boldsymbol{\phi}_4 = \boldsymbol{\phi}_5 = (1,1,3)$, respectively.
In this example, we have two max-min fair allocations: $\pi_4$ and $\pi_5$, because their allocation vectors $\boldsymbol{\phi}_4$ and $\boldsymbol{\phi}_5$, respectively, are lexicographically greater than any other vectors derived from $\pi_1$, $\pi_2$, and $\pi_3$.

Concerning the third objective of the allocation, we notice that $\pi_4$ has a total compensation of $125$, while $\pi_5$ has a total compensation of $105$. Thus, in this example, the optimal allocation that satisfies all three objectives is $\pi_5$ as it has the optimal max-min fairness with the least compensation.$\blacksquare$
\end{example}

We now formally define the optimization problem that we study in this paper.

\begin{definition}[Max-min fair minimum cost allocation (MFMCA) problem]
Given a set of available jobs $\mathcal{J}$ with their possible starting times, suppose a set of valid bids $\mathcal{B}=\{B_1,\ldots,B_K\}$ is submitted by $K$ bidders. Each bid $B_k=\langle \boldsymbol{c}_k, \boldsymbol{n}_k \rangle$ specifies for each bid job $j_i$ its starting time and the desired compensation $c(j^t_i,k)$, together with the company's capacity $n^t_k$ for each time period $t\in \mathcal{T}$.
The objective of the max-min fair minimum cost allocation problem is to find the optimal feasible allocation $\pi_{\boldsymbol{\phi}_f}: \mathcal{J}\times \mathcal{T}  \times \mathcal{K} \mapsto \{0,1\}$, such that the number of allocated jobs $Z=\card{\{j_i: j_i\in \mathcal{J}, \pi(j_i,\cdot,\cdot)=1\}}$ is maximum, and the allocation leads to a max-min fairness vector $\boldsymbol{\phi}_f$ with regard to $Z$ jobs, with the least total compensation $\sum_{j\in \mathcal{J}, k\in \mathcal{K}, t\in \mathcal{T}, \pi_{\boldsymbol{\phi}_f}(j,t,k) =1}c(j^t_i,k)$.
\end{definition}

We treat MFMCA as a two-level optimization problem.
First, we determine what allocation is deemed max-min fair, and second, we determine which of the possibly many max-min fair allocations has the lowest cost.

\begin{definition}[Max-min fair allocation (MMFA) problem]
The objective of the max-min fair allocation problem
is to find the optimal max-min fairness vector $\boldsymbol{\phi}_f$ that indicates the maximum number of jobs that can be assigned feasibly and that leads to a max-min fair allocation among all bidders.
\end{definition}

Given the output of the first-level optimization problem (MMFA), i.e., a max-min fairness vector, we search for the allocation that gives the desired fair allocation and that has the lowest total compensation.

\section{Polynomial-time optimal algorithm for MFMCA}\label{sec:opt}
In this section, we introduce a two-stage network flow based polynomial-time algorithm to solve the proposed MFMCA problem. In the first stage, we propose an iterative maximum flow algorithm, called IMaxFlow, to enforce a fairest job distribution over companies while ensuring that the maximal number of jobs can be allocated. The output of the IMaxFlow algorithm, i.e., the optimal max-min fairness vector $\boldsymbol{\phi}_f$, is then used as input to the FairMinCost algorithm to construct a new flow network. By any standard minimum-cost maximum-flow algorithm on this constructed flow network, we prove that we obtain the optimal solution to MFMCA. In the next section we present the proposed two-stage algorithm, starting with the iterative maximum flow (IMaxFlow) algorithm.

\subsection{IMaxFlow algorithm for solving MMFA}
Given an instance of the MMFA problem, we can construct a network flow, and then apply the proposed iterative maximum flow algorithm to obtain the optimal max-min fairness vector.

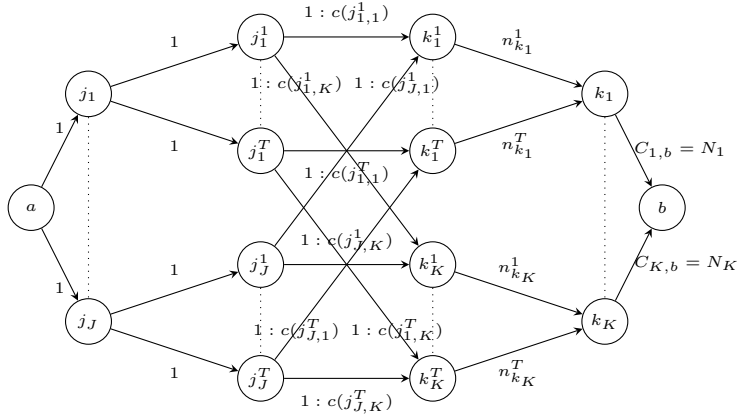
\begin{figure}
\centering
\resizebox{10cm}{!}{
{\scriptsize
\begin{tikzpicture}[>=stealth, scale=0.75,every node/.style={minimum size=0.6cm, inner sep=0pt, font=\tiny}]
	\node [circle,draw] (source) at (0,0) {$a$};
	\node [circle,draw] (j1) at (1,2) {$j_1$};
	\node [circle,draw] (jm) at (1,-2) {$j_J$};
    \draw[->] (source) edge node [above]{1} (j1);
    \draw[->] (source) edge node [below]{1} (jm);
	\draw[dotted] (j1) edge node {} (jm);
	\node [circle,draw] (j1t1) at (4,3) {$j_1^1$};
	\node [circle,draw] (j1tT) at (4,1) {$j_1^T$};
	\node [circle,draw] (jmt1) at (4,-1) {$j_J^1$};
	\node [circle,draw] (jmtT) at (4,-3) {$j_J^T$};
    \draw[->] (j1) edge node [above] {1} (j1t1);
    \draw[->] (j1) edge node [below] {1} (j1tT);
	\draw[dotted] (j1t1) edge node {} (j1tT);
    \draw[->] (jm) edge node [above] {1} (jmt1);
    \draw[->] (jm) edge node [below] {1} (jmtT);
	\draw[dotted] (jmt1) edge node {} (jmtT);
	\node [circle,draw] (t1k1) at (7,3) {$k_1^1$};
	\node [circle,draw] (tTk1) at (7,1) {$k_1^T$};
	\node [circle,draw] (t1knC) at (7,-1) {$k_K^1$};
	\node [circle,draw] (tTknC) at (7,-3) {$k_K^T$};
	\draw[->] (j1t1) edge node [above] {$1:c(j_{1,1}^1)$} (t1k1);
    \draw[->] (j1t1) edge node [pos=0.15] {$1:c(j_{1,K}^1)$} (t1knC);
    \draw[->] (j1tT) edge node [below] {$1:c(j_{1,1}^T)$} (tTk1);
    \draw[->] (j1tT) edge node [pos=0.85] {$1:c(j_{1,K}^T)$} (tTknC);
    \draw[->] (jmt1) edge node [pos=0.85] {$1:c(j_{J,1}^1)$} (t1k1);
    \draw[->] (jmt1) edge node [above] {$1:c(j_{J,K}^1)$} (t1knC);
    \draw[->] (jmtT) edge node [pos=0.15] {$1:c(j_{J,1}^T)$} (tTk1);
    \draw[->] (jmtT) edge node [below] {$1:c(j_{J,K}^T)$} (tTknC);
	\draw[dotted] (t1k1) edge node {} (tTk1);
	\draw[dotted] (t1knC) edge node {} (tTknC);
	\node [circle,draw] (k1) at (10,2) {$k_1$};
	\node [circle,draw] (knC) at (10,-2) {$k_K$};
	\draw[->] (t1k1) edge node [above] {$n_{k_1}^1$} (k1);
	\draw[->] (tTk1) edge node [below] {$n_{k_1}^T$} (k1);
	\draw[->] (t1knC) edge node [above] {$n_{k_K}^1$} (knC);
	\draw[->] (tTknC) edge node [below] {$n_{k_K}^T$} (knC);
	\draw[dotted] (k1) edge node {} (knC);
	\node [circle,draw] (sink) at (11,0) {$b$};
	\draw[->] (k1) edge node [right] {$C_{1,b}=N_{1}$} (sink);
    \draw[->] (knC) edge node [right] {$C_{K,b}=N_{K}$} (sink);
\end{tikzpicture}
}
}
\caption{A constructed flow network for solving MMFA, where $j_1,\ldots,j_J$ represent a set of available jobs, $j^1_1,\ldots,j^T_J$ are job-time nodes, $k^1_1,\ldots,k^T_K$ are company-time nodes, and $k_1,\ldots,k_K$ represent a set of companies.
\label{fig:mlmf}}
{}
\end{figure}

Suppose the set of available jobs is $\mathcal{J}$. We want to build a flow network to push $\mathcal{J}$ from the source node $a$ to the sink node $b$. The flow network is a directed graph  $G = (V,A)$ with capacities $C_{u,v}$ for each $(u,v) \in A$. The flow network can be constructed from any problem instance of MMFA by adding the following node layers and arcs from $a$ to $b$: (1) First, we create a node layer for the jobs $\mathcal{J}$. Each job $j_i \in \mathcal{J}$ of this job layer is connected with source node $a$. Because each job only needs to be executed once, the capacity of these arcs is $1$.
(2) As each job has certain time periods in which it can be executed, we construct another node layer next to the job layer with job-time nodes $j_i^t$ for each available time period $t$ for each job $j_i$. The job-time nodes are connected to their corresponding job nodes in the job layer with the arcs having a capacity equal to $1$, because a job can only be executed at most once in a certain time period.
(3) From the bids of the companies we know which companies bid on which jobs at which time periods with a certain cost. Therefore, from these bids we can construct yet another node layer with company-time nodes that indicate the time periods $t$ in which each company $k$ is available, denoted by $k^t$. These nodes are connected to the corresponding job-time nodes where the company made a bid at that particular time period. These arcs each have a capacity of $1$.
However, unlike previously created arcs, these arcs have costs associated with them equal to the corresponding compensations indicated in the bids. These costs do not play a role in solving MMFA, as its objective is not related to the cost.
(4) Once we have constructed this company-time layer, we can construct another node layer consisting of company nodes. Each node in this company layer corresponds to a company $k \in \mathcal{K}$. The company-time nodes in the company-time layer will then be connected to their respective companies in the company layer to aggregate the former. These arcs have a capacity $n^t_k$ equal to the capacity that a company $k$ has indicated as being available in that particular time period $t$.
Finally, we connect all nodes in the company layer with sink node $b$. For each company $k \in \mathcal{K}$ the edge between its node and the sink has a capacity $N_{k}=\sum_{t\in \mathcal{T}}n^t_{k}$, which is the total capacity over all time slots.
An example of the resulting flow network is illustrated in Figure \ref{fig:mlmf}.

Given the constructed flow network $G$, the value of a flow $f=f(a,b)$ is the total flow that can be pushed from the nodes in the company layer to the sink node $b$, i.e., $f=\sum_{v\in \{k_1,\ldots,k_{K}\}}f(v,b)$. Hence, it is clear that the solution to the problem of finding the maximum flow given the translated flow network problem is equivalent to finding the maximum number of jobs that can be allocated to the companies in MMFA. Therefore, given an instance of the MMFA problem, if we run a standard maximum flow algorithm on the constructed flow network $G$, we will obtain a solution that tells us the maximum number of jobs that can be allocated.

However, the objective of the MMFA problem is also to find the optimal max-min fair solution.
Therefore, to solve this maximum flow problem with an additional fairness property, we introduce an iterative maximum flow algorithm that applies the maximum flow algorithm, such as Edmonds-Karp~\citep{edmonds1972theoretical}, in a greedy fashion. In this way, the flow assigned to each company is increased step by step until no more flows can be assigned. This idea is similar to the so-called progressive filling algorithm \citep{bertsekas1987data}.
Our proposed iterative algorithm IMaxFlow works as follows.

\paragraph{\textbf{Initiation}}
We construct a set $\mathcal{Q}$ that contains all companies and we set the capacity for all companies to 0, which means that in $G$, the capacity on the arcs connecting the nodes $k_1,\ldots, k_{K} \in \mathcal{K}$ in the company layer and the sink node $b$ is set to $0$, i.e., $C_{k,b}=0$ for all $k\in \mathcal{K}$ in Figure~\ref{fig:mlmf}.

\paragraph{\textbf{Iterations}} In the first iteration $I_1$, we arbitrarily pick a company $k_q \in \mathcal{Q}$ and then increase its capacity by $1$, i.e., $C_{k_q,b}=C_{k_q,b}+1=1$. We then run a standard maximum flow algorithm, which returns a maximum flow $f^{I_1}_{k_q}$ given the restricted capacities. We check whether $k_q$ receives a flow, i.e., whether $f^{I_1}(k_q,b)=1$ is true. If $f^{I_1}(k_q,b)=0$, then we can conclude that company $k_q$ will not be allocated any job even if we would further increase its capacity. In this case, we fix the capacity $C^f_{k_q,b}$ of the edge between the sink and company $k_q$ to $C^f_{k_q,b}=1-1=0$, and remove company $k_q$ from set $\mathcal{Q}$.
If $f^{I_1}(k_q,b)=1$, then we know that company $k_q$ can handle a flow of $1$, so we can let $C_{k_q,b}=1$ and continue.
We then choose another company in $\mathcal{Q}$ and repeat the above-mentioned process until we have done the same for all companies in $\mathcal{Q}$. Recall that all companies can get at most one job in this iteration because their capacity is set to $1$.

We then start the next iteration $I_2$. We arbitrarily pick a company $k_q \in \mathcal{Q}$ and check whether it has reached its total capacity. If so, we fix its capacity $C^f_{k_q,b}=C_{k_q,b}$ and remove $k_q$ from $\mathcal{Q}$. Otherwise, we increase its capacity to $C_{k_q,b}=C_{k_q,b}+1=2$.
We again run the maximum flow algorithm on $G$ with the updated capacity and obtain a maximum flow $f^{I_2}_{k_q}$. If the maximum flow $f^{I_2}_{k_q}$ is the same as the maximum flow obtained in the previous step (for the first company in iteration $I_2$, this is the flow at the end of the previous iteration, $f^{I_1}$), we can conclude that increasing the capacity $C_{k_q,b}$ for company $k_q$ does not result in a larger flow. Therefore, we fix the capacity $C^f_{k_q,b}$ of the edge between the sink and company $k_q$ to $C^f_{k_q,b}=2-1=1$, and remove company $k_q$ from $\mathcal{Q}$. We repeat this for all other companies $k_q \in \mathcal{Q}$. For the subsequent companies in the same iteration, we compare the flow obtained after running the maximum flow algorithm on $G$ with the maximum flow obtained in the previous step, which is $f^{I_2}_{k_{q-1}}$.
If the maximum flow $f^{I_2}_{k_q}$ is larger than the maximum flow obtained in the previous step, then we can let $C_{k_q,b}=2$ and continue.

In this way, during iteration $I_i$ we fix a company $k_q$'s capacity to $C^f_{k_q,b}=i-1$ in $G$, either when the company does not receive more flow than in the previous step $I_{i,k_{q-1}}$ (or $I_{i-1}$ if $k_q$ is first in $I_{i}$), or when the company reaches its maximal total capacity, i.e., $C^f_{k_q,b}=i-1=N_{k_q}$. In each iteration we always add one more capacity to the company-sink edges whose capacities have not been fixed.

\paragraph{\textbf{Termination}}
We iterate this process until $\mathcal{Q}$ is empty, that is, when the flow no longer increases with the addition of more capacities to the companies, or when the capacities of all the companies have reached their limits. It also follows that the capacities of all the company-sink arcs are fixed to some values.

We return the maximum flow $f$ found upon termination as the maximum number of jobs that can be allocated,
and the fixed capacities $C^f_{k,b}$. The fixed capacities $C^f_{k,b}$ --- equivalent to the number of flows on the company-sink edges, $f(k_1,b)$, $f(k_2,b)$, $\ldots$, $f(k_{K},b)$ --- specifies the number of jobs $\omega_{k_1}$, $\omega_{k_2}$, $\ldots$, $\omega_{k_{K}}$ assigned to companies $k_1, \ldots, k_{K}$. The fixed capacities $C^f_{k,b}$ also comprise the max-min fairness vector $\boldsymbol{\phi}_f= \sigma(\boldsymbol{\omega})$.

\paragraph*{}
This iterative maximum flow algorithm is described in Algorithm~\ref{alg_mlmf}. Note that this adaptation is independent of the maximum flow algorithm used and is therefore suitable to be used in combination with any existing maximum flow algorithm.

\begin{algorithm}
\renewcommand{\algorithmicrequire}{\textbf{Input:}}
\renewcommand{\algorithmicensure}{\textbf{Output:}}
\caption{IMaxFlow algorithm for solving MMFA}
\label{alg_mlmf}
\begin{algorithmic}
\REQUIRE $G = (V,A)$ a constructed flow network for an instance of MMFA, where $a,b$ are the source and sink node, respectively. The capacity of a company-sink edge is denoted as $C_{k,b}$ for $k\in \mathcal{K}$. $N_k$ denotes the maximum capacity of company $k$
\ENSURE a maximum flow $f$ and a max-min fair allocation vector $\boldsymbol{\phi}_f$
\STATE $f_{curr} \gets 0 $; $f_{prev} \gets -1$
\STATE $\mathcal{Q} = \mathcal{K}$; $I=0$~~\COMMENT{$I$ denotes the iteration number} 
\STATE $C^f_{k,b} \leftarrow 0$, $\forall~ k\in \mathcal{K}$ ~~\COMMENT{$C^f_{k,b}$ denotes the final fixed capacity for company-sink edge $e(k,b)$}
\STATE $C_{k,b} \gets 0$, $\forall~ k\in \mathcal{K}$ ~~\COMMENT{update $G$ by setting capacities of company-sink edges to 0}
\WHILE{$\mathcal{Q} \neq \emptyset$}
\STATE $I = I + 1$ ~~~~\COMMENT{increase the iteration number by 1}
\FOR{each $k \in \mathcal{Q}$}
\STATE $f_{prev} \gets f_{curr}$
\IF{$C_{k,b} < N_{k}$}
\STATE $C_{k,b} \gets C_{k,b} + 1$
\STATE Call maximum flow algorithm (MF) on $G$, $f_{curr} \gets$ MF$(G)$
\IF{$f_{curr} = f_{prev}$}
\STATE $C_{k,b} \gets C_{k,b} - 1$; $C^f_{k,b} \gets C_{k,b}$
\STATE $\mathcal{Q} \gets \mathcal{Q} \setminus \{k\}$
\ENDIF
\ELSE
\STATE $C^f_{k,b} \gets C_{k,b}$, $\mathcal{Q} \gets \mathcal{Q} \setminus \{k\}$
\ENDIF
\ENDFOR
\ENDWHILE
\RETURN $f_{curr}$ as $f$, sorted $(C^f_{1,b},\ldots, C^f_{K,b})$ as 
$\boldsymbol{\phi}_f$
\end{algorithmic}
\end{algorithm}

We illustrate IMaxFlow by the following example.

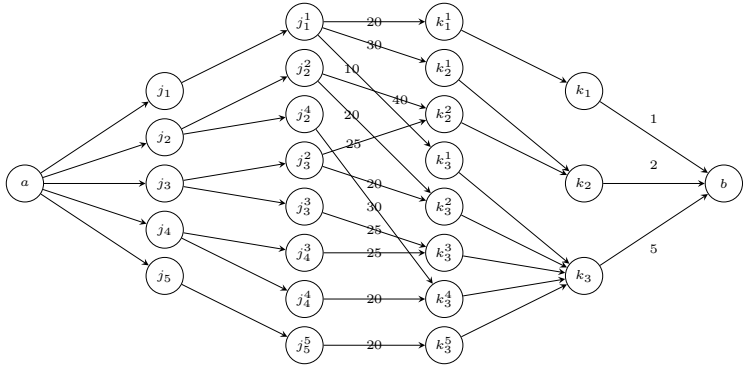
\begin{figure}
\centering
\resizebox{10cm}{!}{
{\scriptsize
\begin{tikzpicture}[>=stealth, scale=0.75,every node/.style={minimum size=0.6cm, inner sep=0pt, font=\tiny}]
		\node [circle,draw] (source) at (-2,0) {$a$};
		\node [circle,draw] (j1) at (1,2) {$j_1$};
		\node [circle,draw] (j2) at (1,1) {$j_2$};
		\node [circle,draw] (j3) at (1,0) {$j_3$};
		\node [circle,draw] (j4) at (1,-1) {$j_4$};
		\node [circle,draw] (j5) at (1,-2) {$j_5$};
	    \draw[->] (source) edge node [below]{} (j1);
	    \draw[->] (source) edge node [below]{} (j2);
	    \draw[->] (source) edge node [below]{} (j3);
	    \draw[->] (source) edge node [below]{} (j4);
	    \draw[->] (source) edge node [below]{} (j5);
		\node [circle,draw] (j1t1) at (4,3.5) {$j_1^1$};
		\node [circle,draw] (j2t2) at (4,2.5) {$j_2^2$};
		\node [circle,draw] (j2t4) at (4,1.5) {$j_2^4$};
		\node [circle,draw] (j3t2) at (4,0.5) {$j_3^2$};
		\node [circle,draw] (j3t3) at (4,-0.5) {$j_3^3$};
		\node [circle,draw] (j4t3) at (4,-1.5) {$j_4^3$};
		\node [circle,draw] (j4t4) at (4,-2.5) {$j_4^4$};
		\node [circle,draw] (j5t5) at (4,-3.5) {$j_5^5$};
	    \draw[->] (j1) edge node [below] {} (j1t1);
		\draw[->] (j2) edge node [below] {} (j2t2);
		\draw[->] (j2) edge node [below] {} (j2t4);
	    \draw[->] (j3) edge node [below] {} (j3t2);
	    \draw[->] (j3) edge node [below] {} (j3t3);	
	    \draw[->] (j4) edge node [below] {} (j4t3);
	    \draw[->] (j4) edge node [below] {} (j4t4);
	    \draw[->] (j5) edge node [below] {} (j5t5);
		\node [circle,draw] (t1k1) at (7,3.5) {$k_1^1$};
		\node [circle,draw] (t1k2) at (7,2.5) {$k_2^1$};
		\node [circle,draw] (t2k2) at (7,1.5) {$k_2^2$};
		\node [circle,draw] (t1k3) at (7,0.5) {$k_3^1$};
		\node [circle,draw] (t2k3) at (7,-0.5) {$k_3^2$};
		\node [circle,draw] (t3k3) at (7,-1.5) {$k_3^3$};
		\node [circle,draw] (t4k3) at (7,-2.5) {$k_3^4$};
		\node [circle,draw] (t5k3) at (7,-3.5) {$k_3^5$};
		\draw[->] (j1t1) edge node [pos=0.5] {20} (t1k1);
		\draw[->] (j1t1) edge node [pos=0.5] {30} (t1k2);
		\draw[->] (j1t1) edge node [pos=0.3] {10} (t1k3);
	    \draw[->] (j2t2) edge node [pos=0.75] {40} (t2k2);
	    \draw[->] (j2t2) edge node [pos=0.3] {20} (t2k3);
	    \draw[->] (j2t4) edge node [pos=0.5] {30} (t4k3);
	    \draw[->] (j3t2) edge node [pos=0.3] {25} (t2k2);
		\draw[->] (j3t2) edge node [pos=0.5] {20} (t2k3);
		\draw[->] (j3t3) edge node [pos=0.5] {25} (t3k3);
	    \draw[->] (j4t3) edge node [pos=0.5] {25} (t3k3);
	    \draw[->] (j4t4) edge node [pos=0.5] {20} (t4k3);
	    \draw[->] (j5t5) edge node [pos=0.5] {20} (t5k3);
		\node [circle,draw] (k1) at (10,2) {$k_1$};
		\node [circle,draw] (k2) at (10,0) {$k_2$};
		\node [circle,draw] (k3) at (10,-2) {$k_3$};
		\draw[->] (t1k1) edge node [below] {} (k1);
		\draw[->] (t1k2) edge node [below] {} (k2);
		\draw[->] (t2k2) edge node [below] {} (k2);
		\draw[->] (t1k3) edge node [below] {} (k3);
		\draw[->] (t2k3) edge node [below] {} (k3);
		\draw[->] (t3k3) edge node [below] {} (k3);
		\draw[->] (t4k3) edge node [below] {} (k3);
		\draw[->] (t5k3) edge node [below] {} (k3);
		\node [circle,draw] (sink) at (13,0) {$b$};
		\draw[->] (k1) edge node [above] {1} (sink);
		\draw[->] (k2) edge node [above] {2} (sink);
		\draw[->] (k3) edge node [below] {5} (sink);
\end{tikzpicture}
}
}
\caption{The constructed flow network given the problem instance described in Example \ref{example_MLFA}. The capacities of the arcs in the flow network are 1, except for the arcs between the company nodes $k_k$ and the sink $b$. The numbers on the edges between the job-time nodes $j^t_i$ and the company-time nodes $k^t_k$ specify the compensations of company $k$ performing job $j_i$ at time period $t$. We do not take these costs into account in MMFA.
\label{fig:example_flow}}
{}
\end{figure}

\begin{example}\label{example_flow}
Refer to the problem instance in Example \ref{example_MLFA}. We can construct the accompanying flow network as shown in Figure \ref{fig:example_flow}.
The IMaxFlow algorithm first sets all the capacities of the three companies --- i.e., the edges $e(k,b)$ connecting to sink $b$ --- to $0$. Then it increases the capacity of $e(k_1,b)$ by $1$ and runs the maximum flow algorithm, which obtains $f(k_1,b)=1$. This is repeated for each company.  At the end of the first iteration we have $f^{I_1}(k_1,b)=f^{I_1}(k_2,b)=f^{I_1}(k_3,b)=1$, and the total maximum flow is $f^{I_1} = 3$. This can be achieved by pushing a flow from $j_1$ to $k_1$, a flow from $j_2$ to $k_2$, and a flow from $j_3$ to $k_3$.

During the second iteration, $C^f_{k_1,b}$ is fixed to $1$ as $k_1$ has reached its highest capacity and $f^{I_2}_{k_1} = f^{I_1} = 3$.
Next, the capacity of $e(k_2,b)$ is set to $2$. After running a standard maximum flow algorithm, we have a maximum flow $f^{I_2}_{k_2}=3$, because $k_1$ and $k_2$ together can be assigned two jobs (either $j_1, j_2$ or $j_1, j_3$) and $k_3$ receives one job because its capacity is still $1$. As $f^{I_2}_{k_2}=f^{I_2}_{k_1}$, increasing $k_2$'s capacity does not help to increase the flow but may harm the fairness because it may happen that both $j_1, j_2$ (or  $j_1, j_3$) can be allocated to $k_2$. Hence, we fix $k_2$'s capacity $C^f_{k_2,b}$ to $1$.
We then look at the case where the capacity of $e(k_3,b)$ is increased to $2$.  It is clear that $f^{I_2}_{k_3}$ is now $4$.

Thus, we continue with iteration 3, where we only increase $k_3 $'s capacity to $3$. After running IMaxFlow, we have a flow of $5$, with a possible allocation of $j_1$ to $k_1$, $j_3$ to $k_2$, and $j_2$, $j_4$, $j_5$ to $k_3$.

As increasing $k_3$'s capacity will not increase the flow any further, $C^f_{k_3,b}$ is fixed to $3$, and
the algorithm terminates.
The maximum number of allocated jobs is $5$, with a max-min fairness vector of $\boldsymbol{\phi}_f = (1,1,3)$, which is simply the fixed capacity of each company-sink edge sorted in nondecreasing order.
$\blacksquare$
\end{example}

We now prove that IMaxFlow is correct, that is, the returned flow value $f$ is the maximum number of jobs that can be allocated, and the returned fairness vector $\boldsymbol{\phi}_f$ is the most fair job distribution over the participating companies given $f$.  

\begin{theorem}\label{thm:imaxflow}
IMaxFlow allocates the maximum number of jobs to the companies and returns a sorted allocation vector that is max-min fair.
\end{theorem}

\begin{proof}

We will prove by induction that given a set of companies $\mathcal{K}$, at any iteration $I_i$ of the algorithm IMaxFlow, given the available capacities of $\mathcal{K}$,  the returned flow $f^{I_i}$ is maximum, and the sorted allocation vector is max-min fair among all $f^{I_i}$-feasible vectors.

\noindent {\bf Base case:} All companies start with capacity $0$. In the first iteration $I_1$ of IMaxFlow, one company $k\in \mathcal{K}$ is arbitrarily picked and assigned a capacity of $1$. 
Then we run the maximum flow (MF) algorithm, which determines the maximum flow of the network given the current available capacity. If this added capacity did not increase the total flow,  $k$'s capacity is fixed to $0$.
At the end of iteration $I_1$, when the last company is given a capacity of $1$ and the maximum flow algorithm is run, it is clear that the returned flow $f^{I_1}$ is maximum given the total capacity of $\mathcal{K}$. 
Let $\mathcal{K}_0$ be the set of companies whose capacities have been fixed to 0 during this iteration. Then the sorted allocation vector is
\[
 \boldsymbol{\phi}_{f^{I_1}}= ( \underbrace{0, \ldots, 0,}_{\card{\mathcal{K}_0}}  \underbrace{1,\ldots,1}_{K-\card{\mathcal{K}_0}} ).
 \]
It is possible that the set $\mathcal{K}_0$ is not unique. For example, a flow can be pushed either through $j$'s node or $k$'s node. If we pick $j$ first to increase the capacity and to test the flow, then later increasing $k$'s capacity to 1 will not increase the total flow and hence $k$'s capacity will be fixed to 0, i.e., $k\in \mathcal{K}_0$. On the other hand, if we pick $k$ earlier than $j$, we will have $j \in \mathcal{K}_0$. This situation however gives us the same sorted vector of two companies, which is $(0,1)$. Thus,
the first iteration of the algorithm may result in a different set  $\mathcal{K}_0$, but the size of $\mathcal{K}_0$ is always the same and  the total number of flows $f^{I_1}$ that can be pushed is always maximum. Therefore, the resulting sorted allocation vector is the same for all possible $f^{I_1}$-feasible vectors, and it is max-min fair.
Thus the statement holds for the first iteration $I_1$.

\noindent {\bf Induction step:} Suppose the statement is true for iteration $I_i$, that is, after this iteration, the returned flow $f^{I_i}$ is maximum given the total capacity added, and the sorted allocation vector $\boldsymbol{\phi}_{f^{I_i}}$ is max-min fair.  
Let $\boldsymbol{\phi}_{f^{I_i}}$ be
\[
 \boldsymbol{\phi}_{f^{I_i}}= ( \underbrace{C^f_{1,b}, \ldots, C^f_{m,b},}_{ \text{fixed}}  \underbrace{i,\ldots,i}_{K-m} ).
 \]
In $\boldsymbol{\phi}_{f^{I_i}}$, suppose there are $m$ company-sink edges with fixed capacities $C^f_{h,b}$, $1 \leq h \leq m$. 
We denote these companies as $\mathcal{K}_{\text{fix}}$. For the remaining unfixed $K-m$ company-sink edges, according to the algorithm, the amount of flow must be equal to their assigned capacity on iteration $I_i$, which is $i$. 
Hence, we have for $I_i$ the maximum flow $f^{I_i}=\sum_{ h \in \mathcal{K}_{\text{fix}} }C^f_{h,b} + i \times (K-m)$.

Now we need to show the statement stays true for iteration $I_{i+1}$. During this iteration, each company $j \notin \mathcal{K}_{\text{fix}}$, who does not have a fixed capacity, is assigned one more capacity to have a total capacity of $i+1$. Let $j \notin \mathcal{K}_{\text{fix}}$ be the first company to increase the capacity.
After running the MF algorithm, the returned maximum flow is either $f^{I_i}$ or $f^{I_i}+1$, corresponding to the cases that
$j$ will receive either $i$ flow or $i+1$ with an extra capacity.  If $j$ receives $i$ flow, it is because
either its original capacity $N_j=i$ or only $i$ flow can be pushed along the job nodes to company $j$'s node.
At the end of iteration $I_{i+1}$, all companies not in $\mathcal{K}_{\text{fix}}$ have been given one more capacity and have been tested with the MF algorithm. 
Assume $L$ companies not in $\mathcal{K}_{\text{fix}}$ will be assigned $i+1$ flows after iteration $I_{i+1}$. Then the total flow $f^{I_{i+1}}=f^{I_{i}}+L$ is maximum given the capacity of this iteration, as $f^{I_{i}}$ is maximum at iteration $I_i$. The resulting sorted allocation vector is
\[
 \boldsymbol{\phi}_{f^{I_{i+1}}}= ( \underbrace{C^f_{1,b}, \ldots, C^f_{m,b},}_{m}  \underbrace{i,\ldots,i,}_{K-m-L} \underbrace{i+1,\ldots,i+1}_{L}).
 \]
Similar to the reasoning for the base case of iteration $I_1$, these $L$ companies could be different depending on the ordering of adding one extra capacity and testing. However, the sorted allocation vector for the companies in $\mathcal{K}_{\text{fix}}$ is always the same, i.e, $(\underbrace{i,\ldots,i,}_{K-m-L} \underbrace{i+1,\ldots,i+1}_{L})$. Together with the fact that $\boldsymbol{\phi}_{f^{I_i}}$ is max-min fair in the previous iteration $I_i$, we have shown that $\boldsymbol{\phi}_{f^{I_{i+1}}}$ is max-min fair among $f^{I_{i+1}}$-feasible allocation vectors.

\noindent {\bf Conclusion:} By the principle of induction, it follows that the preceding statement is true for any iteration of the algorithm IMaxFlow.

Hence, it follows that after the final iteration IMaxFlow returns the maximum number of jobs to the companies and the sorted allocation vector is max-min fair.

\end{proof}

As a by-product of the above reasoning, we have the following lemma.

\begin{lemma}
IMaxFlow returns a unique max-min fair allocation vector, given the maximum number of allocated jobs.
\end{lemma}

Finally, we show that the proposed algorithm is a polynomial-time algorithm (see \ref{app:runtime1} for the proof).

\begin{theorem}\label{lemma_mlmf_runningtime}
The IMaxFlow algorithm runs in time $O((J^3K^3T^3) + (J^2K^4T^3))$.
\end{theorem}



\subsection{FairMinCost algorithm}\label{sec:mfmca}
Once we know the fairness vector from IMaxFlow, we want to minimize the associated cost (compensations). Because there are many feasible max-min fair maximum flow solutions with different costs, we want to find the one with the minimum cost. Unfortunately, we cannot apply a standard minimum-cost maximum-flow algorithm to our flow network as it may violate the max-min fairness condition while looking for the minimum cost.
The obtained fairness vector tells us in what quantities the jobs will be distributed in the fairest allocation.
However, we do not know \emph{which} company would be assigned \emph{which} number of jobs such that the total cost is smallest.

If we know the exact number of jobs all companies would get, 
MFMCA is easily solvable using a minimum-cost maximum-flow algorithm.
This is obvious because we can set the capacities of the arcs from the company nodes to the sink to be equal to the number of jobs of the respective companies. Since we know from MMFA that the flow is maximal and feasible, and that the capacities sum up to this maximum flow, we know that all jobs will be assigned. This boils down to a simple minimum-cost maximum-flow problem that can be solved using any of the existing algorithms.

However, if the exact number of jobs that all companies will get is not known, then the capacity for each company can be any of the capacities in the fairness vector. This leaves us with many ways to construct the flow network because it is assumed that the capacity of the arcs in a minimum-cost maximum-flow problem are known. We can deal with this problem in several ways.

One way to find the minimum cost among all possible max-min fair allocations is to simply enumerate all possible max-min fair allocations and solve a minimum-cost maximum-flow problem for each of them, and then to finally choose the allocation that has minimum cost.
However, this method would be computationally inefficient, because it can be viewed as a multiset permutation with $\binom{p}{r_1,r_2,\dots,r_{p}} = \frac{p!}{r_1!r_2! \dots r_{p}!}$ possibilities,
where $p = \card{\boldsymbol{\phi}_f^U}$, in which $\boldsymbol{\phi}_f^U$ denotes the vector of unique capacities in $\boldsymbol{\phi}_f$, and $r_i$ denotes how often capacity $i$ appears in $\boldsymbol{\phi}_f$, $r_i = \sum_{k \in \mathcal{K}} \card{\boldsymbol{\phi}_f(k) = i}$. For each possibility, we would need to run a minimum-cost maximum-flow algorithm. The resulting running time would be exponential.

Instead, in this paper we propose an algorithm that makes variable capacities on arcs in the flow network possible. Given the fairness vector $\boldsymbol{\phi}_f=(\phi_1,\ldots,\phi_{K})$,
we introduce a solution method that runs in polynomial time. To this end, we adjust the network flow model such that the fair job distribution $(\phi_1,\ldots,\phi_{K})$ will be intact
at the same time that cost minimization takes place. The challenge is to somehow enforce the capacities of the fairness vector obtained from IMaxFlow in the final allocation.

For ease of explanation, we denote an instance of the original MFMCA problem as $P=\langle \mathcal{J},\mathcal{K}, \mathcal{T}, \mathcal{B} \rangle$.
We now introduce a new problem $P'=\langle \mathcal{J}', \mathcal{K}', \mathcal{T}', \mathcal{B}' \rangle$ adapted from $P$.  The key construction of $P'$ given $P$ is that we will update the original set of jobs $\mathcal{J}$ to $\mathcal{J}'=\mathcal{J} \cup \mathcal{J}^d$, where $\mathcal{J}^d$ is a set of dummy jobs. Each dummy job provides a flow of $1$ and has a cost of $0$. These dummy jobs will be performed during dummy time periods $\mathcal{T}^d$, thus, $\mathcal{T}'=\mathcal{T}\cup \mathcal{T}^d$.
The set of companies $\mathcal{K}'$ in $P'$ stays the same as in the original problem $P$, i.e., $\mathcal{K}'=\mathcal{K}$, however, they have capacities at dummy periods $\mathcal{T}^d$ for performing dummy jobs $\mathcal{J}^d$.
The number of dummy jobs and dummy periods will be determined by the fairness vector $\boldsymbol{\phi}_f=(\phi_1,\ldots,\phi_{K})$ returned by the IMaxFlow algorithm on the instance of $P$. We construct a flow network $G'$ for problem $P'$ based on the constructed flow network $G$ for an instance of problem $P$ by adding the dummy jobs and dummy times in $G$. In addition, we update the capacities of all companies to $\phi_{K}$.
After completing $G'$, we claim that if we run a standard minimum cost, maximum flow algorithm on $G'$, the solution is a fair minimum cost job allocation for the original problem $P$. We denote this procedure as the FairMinCost algorithm.

\paragraph{\textbf{Construction of $G'$}.} We first show how we can construct a flow network $G'$ for problem $P'$ based on the constructed flow network $G$ for an instance of problem $P$.
Given $G$, we will add a number of so-called horizontal dummy layers ($DL$ for short). We have a dummy layer for each increment of $1$ from the lowest number of jobs, $\phi_1$, to the highest, $\phi_{K}$. Hence,
the total number of $DL$ is equal to the difference between the lowest and the highest number of jobs in the fairness vector $\boldsymbol{\phi}_f$, i.e., there are $\phi_{K}-\phi_1$ dummy layers: $DL_{1}, DL_{2}, \ldots, DL_{\phi_{K}-\phi_1}$.
Each layer is meant to provide dummy jobs to companies such that all companies can have jobs up to a specified number. We associate each dummy layer to the specified number in $\boldsymbol{\phi}_f$, that is, $DL_{1}$ is associated with number $\phi_1+1$, $DL_{2}$ with $\phi_1+2$, and $DL_{\phi_{K}-\phi_1}$ with $\phi_1+(\phi_{K}-\phi_1)=\phi_{K}$.
In each dummy layer $DL_l$, we create a set of dummy job nodes $\mathcal{J}^d_l$ in the job layer of the network equal to the number of companies that have a lower capacity than $\phi_1+l$ in the fairness vector $\boldsymbol{\phi}_f$. These dummy jobs $d_{l,i} \in \mathcal{J}^d_l$ are connected with source node $a$.
We assume that each dummy layer $DL_l$ has its own unique dummy time $t'_l$ and all dummy jobs from $DL_l$ need to be executed at time $t'_l$.
Thereafter, we create dummy job-time nodes $d^{t'_l}_{l,i}$ in the job-time layer for each dummy job $d_{l,i} \in \mathcal{J}^d_l$, and connect them with their corresponding dummy job nodes $d_{l,i}$.
We assume that every company in $\mathcal{K}'$ is capable of performing every dummy job in $\mathcal{J}^d$, but that for each dummy time $t'_l$ the capacity of every company is $1$.
Thus, for each dummy layer $DL_l$, we create dummy time-company nodes $k^{t'_l}$ in the time-company layer for each company $k \in \mathcal{K}'$ and connect them with all dummy job-time nodes in that particular dummy layer $DL_l$.
Finally, we connect the dummy time-company nodes $k^{t_l}$  to the company nodes $k$ in the company layer of the flow network $G$.
All added arcs have a cost of $0$ and a capacity of $1$. Finally, we change the capacities of the arcs from the company nodes $k$ to the sink $b$ in $G$ to the largest number according to the fairness vector, i.e., $\phi_{K}$.

By creating nodes for each company per dummy layer, we are making sure that each company can be assigned at most one dummy job in each dummy layer. Therefore, each company is able to get any of the capacities in the fairness vector and it is not predetermined which companies are assigned which capacity. This is exactly the flexibility we desire.

\begin{figure}
\centering
\resizebox{10cm}{!}{
{\scriptsize
\begin{tikzpicture}[>=stealth, scale=0.75,every node/.style={minimum size=0.7cm, inner sep=0pt, font=\tiny}]
	\node [circle,draw] (source) at (-2,0) {$a$};
	\node [rectangle, minimum height = 5em, minimum width = 15em,draw] (dotbox) at (4,0) {Partial flow network of $P$ in Figure~\ref{fig:example_flow}};
	\draw[->] (source) edge node [above]{5} (dotbox);
	\node [circle,draw] (k1) at (10,1.5) {$k_1$};
	\node [circle,draw] (k2) at (10,0) {$k_2$};
	\node [circle,draw] (k3) at (10,-1.5) {$k_3$};
	\draw[->] (dotbox) edge node [below] {} (k1);
	\draw[->] (dotbox) edge node [below] {} (k2);
	\draw[->] (dotbox) edge node [below] {} (k3);
	\node [circle,draw] (d1) at (1,-2.5) {$d_{1,1}$};
	\node [circle,draw] (d2) at (1,-3.5) {$d_{1,2}$};
	\node [circle,draw] (d3) at (1,-6) {$d_{2,1}$};
	\node [circle,draw] (d4) at (1,-7) {$d_{2,2}$};
	\draw[->] (source) edge node [above]{1} (d1);
	\draw[->] (source) edge node [below]{1} (d2);
    \draw[->] (source) edge node [above]{1} (d3);
    \draw[->] (source) edge node [below]{1} (d4);
	\node [circle,draw] (d1td1) at (4,-2.5) {$d_{1,1}^{t'_1}$};
	\node [circle,draw] (d2td1) at (4,-3.5) {$d_{1,2}^{t'_1}$};
	\node [circle,draw] (d3td2) at (4,-6) {$d_{2,1}^{t'_2}$};
	\node [circle,draw] (d4td2) at (4,-7) {$d_{2,2}^{t'_2}$};
	\draw[->] (d1) edge node [below]{} (d1td1);
	\draw[->] (d2) edge node [below]{} (d2td1);
    \draw[->] (d3) edge node [below]{} (d3td2);
    \draw[->] (d4) edge node [below]{} (d4td2);
    \node [circle,draw] (td1k1) at (7,-2) {$k_1^{t'_1}$};
	\node [circle,draw] (td1k2) at (7,-3) {$k_2^{t'_1}$};
	\node [circle,draw] (td1k3) at (7,-4) {$k_3^{t'_1}$};
	\draw[->] (d1td1) edge node [below] {} (td1k1);
	\draw[->] (d1td1) edge node [below] {} (td1k2);
	\draw[->] (d1td1) edge node [below] {} (td1k3);
	\draw[->] (d2td1) edge node [below] {} (td1k1);
	\draw[->] (d2td1) edge node [below] {} (td1k2);
	\draw[->] (d2td1) edge node [below] {} (td1k3);
	\node [circle,draw] (td2k1) at (7,-5.5) {$k_1^{t'_2}$};
	\node [circle,draw] (td2k2) at (7,-6.5) {$k_2^{t'_2}$};
	\node [circle,draw] (td2k3) at (7,-7.5) {$k_3^{t'_2}$};
    \draw[->] (d3td2) edge node [below] {} (td2k1);
    \draw[->] (d3td2) edge node [below] {} (td2k2);
    \draw[->] (d3td2) edge node [below] {} (td2k3);
    \draw[->] (d4td2) edge node [below] {} (td2k1);
    \draw[->] (d4td2) edge node [below] {} (td2k2);
    \draw[->] (d4td2) edge node [below] {} (td2k3);
    \draw[->] (td1k1) edge node [below] {} (k1);
	\draw[->] (td1k2) edge node [below] {} (k2);
	\draw[->] (td1k3) edge node [below] {} (k3);
	\draw[->] (td2k1) edge node [below] {} (k1);
	\draw[->] (td2k2) edge node [below] {} (k2);
	\draw[->] (td2k3) edge node [below] {} (k3);
	\node [circle,draw] (sink) at (13,0) {$b$};
	\draw[->] (k1) edge node [above] {3} (sink);
	\draw[->] (k2) edge node [above] {3} (sink);
	\draw[->] (k3) edge node [below] {3} (sink);
\end{tikzpicture}
}
}
\caption{Minimum-cost maximum-flow network with dummy jobs for problem $P'$ for the problem instance described in Example \ref{example_MLFA}. All arcs in the added dummy layers have a cost of $0$ and a capacity of $1$.
\label{fig:dummy}}
{}
\end{figure}
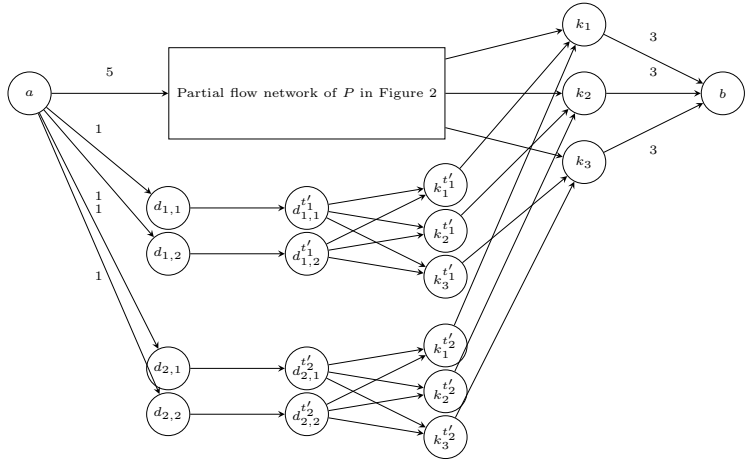

\begin{example}\label{example_dummy}
In Example \ref{example_MLFA}, we obtained a fairness index of $\boldsymbol{\phi}_f = (1,1,3)$, and we showed the constructed flow network $G$ in Figure~\ref{fig:example_flow}. We now show how to add dummy layers to $G$ for this instance in order to obtain $G'$. Figure \ref{fig:dummy} shows the final construction.

Given $\boldsymbol{\phi}_f = (1,1,3)$, we have to create $(3-1)=2$ dummy layers. In the first dummy layer $DL_1$
we create two dummy jobs $d_{1,1}$ and $d_{1,2}$ for the companies that have capacity $1$ in order for each of them to reach a capacity of $2$. We then connect these two dummy jobs to the same dummy time $t'_1$. Thereafter, we create three dummy company-time nodes that are connected to the two dummy job-time nodes and to the three company nodes. The capacity of each arc is $1$. This means that every company is able to do any of the dummy jobs during dummy time $t'_1$ but that only one dummy job from the same dummy layer can be assigned to the same company due to capacity constraints.
Subsequently, we use a similar construction for the second dummy layer $DL_2$. In this layer we again need to create two dummy jobs because there are two capacities smaller than $3$ in $\boldsymbol{\phi}_f$. The cost of all added edges is $0$.
After creating $DL_2$, we change the capacities of the arcs between the companies to the sink node from $(1, 2, 5)$ to $(3, 3, 3)$.
$\blacksquare$
\end{example}

\paragraph{\textbf{Finding minimum-cost maximum-flow}.}
Given the constructed network $G'$, all edges have a cost $\texttt{cost}(e(u,v))$ of 0, except the edges between the original job-time nodes and the original company-time nodes.
We then run any existing (polynomial-time) minimum-cost maximum-flow algorithm on $G'$ that is constructed for problem $P'$. The solution is a flow $f'$ satisfying $f'=\argmin_{u',v'\in V'} \texttt{cost}(e(u,v))f'(u,v)$, and
$f'(k)$ is the number of allocated jobs to company $k \in \mathcal{K}$. Let $\pi'_{k}$ denote the set of jobs assigned to $k$. After removing the dummy jobs in $\pi'_{k}$, i.e., $\pi_{k} = \pi'_{k} \setminus \{d~|~d \in \mathcal{J}^d\}$, we obtain $\pi_{k}$ that is a set of real jobs in $\mathcal{J}$ assigned to company $k$. Then the allocation vector $(\pi_{k_1},\ldots,\pi_{k_K})$ represents an optimal max-min fair allocation with the least costs, which is the solution to the original problem $P$.

To summarize, given an instance $P$ of the problem MFMCA, we use the IMaxFlow algorithm to obtain a max-min fair allocation vector $\boldsymbol{\phi}_f$. Algorithm FairMinCost then constructs an instance $P'$, built upon the flow network of $P$, by adding a set of dummy nodes and arcs determined by $\boldsymbol{\phi}_f$. We then run an existing polynomial-time minimum cost maximum flow algorithm on the flow network of $P'$. In the resulting flow, we remove the dummy jobs and dummy flows.
Our running example demonstrates how the algorithm works.
\begin{example}\label{example_mfmca}
We run an existing minimum cost maximum flow algorithm on the flow network for problem $P'$ (see Figure \ref{fig:dummy}) constructed in Example \ref{example_dummy}. The job allocation $\pi'$ for problem $P'$ is: $k_1$ is assigned $\{j_1^1,d_{1,1},d_{2,1}\}$, $k_2$ is assigned $\{j_3^2,d_{1,2},d_{2,2}\}$, and $k_3$ is assigned $\{j_2^2, j_4^4, j_5^5\}$. All dummy jobs are assigned, and the allocation vector is $(3,3,3)$. The total cost is $105$. We now remove all dummy jobs from $\pi'$ in order to obtain the solution $\pi$ to the original problem $P$. We then have: $j_1^1$ to $k_1$, $j_3^2$ to $k_2$ and $j_2^2, j_4^4, j_5^5$ to $k_3$. The fairness vector of $\pi$ is $\boldsymbol{\phi}_f=(1,1,3)$, which is max-min fair, obtained from the algorithm IMaxFlow as illustrated in Example~\ref{example_flow}. The total cost of $\pi$ is $105$ as dummy jobs have no cost.
This is the same as the optimal max-min fair allocation in Example \ref{example_MLFA}.
$\blacksquare$
\end{example}

We now claim FairMinCost is correct, i.e., the final flow returned by the FairMinCost algorithm gives us an optimal solution to $P$ with a max-min fairness vector $\boldsymbol{\phi}_f$ and has the lowest cost given $\boldsymbol{\phi}_f$.

\begin{theorem}\label{lemma_solution}
The FairMinCost algorithm returns an optimal solution of instance $P$ for MFMCA.
\end{theorem}

The proof is given by the following two lemmas.

\begin{lemma}\label{lemma_fairindex}
The fairness vector obtained from solving problem $P'$ by the FairMinCost algorithm is the max-min fairness vector $\boldsymbol{\phi}_f$ returned as an optimal solution to MMFA.
\end{lemma}

\begin{proof}
Let $\boldsymbol{\phi}_f = (\phi_1, \phi_2, \dots, \phi_{K})$ be the nondecreasingly ordered capacities for the companies in the fairness vector of the optimal solution of MMFA. Hence in an optimal solution to $P$, the allocation vector is also $\boldsymbol{\phi}_f$. In problem $P'$, the capacities are set to $\boldsymbol{\phi}' = (\phi_{K}, \phi_{K}, \dots, \phi_{K})$, as we add $\boldsymbol{\phi}^d = (\phi^d_1, \phi^d_2, \dots, \phi^d_{K}) = (\phi_{K}-\phi_1, \phi_{K}-\phi_2, \dots, \phi_{K}-\phi_{K})$ capacity for the dummy jobs.

We show that in the allocation of the optimal solution of problem $P'$, each company $k \in \mathcal{K}$ will be assigned exactly $\boldsymbol{\phi}_k^d$ dummy jobs by any minimum cost maximum flow algorithm.  
We first note that all dummy jobs will be allocated because there is sufficient capacity added to the network to account for the dummy jobs and they have a cost of zero.

We will show that any company $i \in \mathcal{K}$ cannot get assigned more than $\phi^d_i$ dummy jobs. Let $i=1$ be the first company in the sorted allocation vector, i.e., it has the most dummy capacity. For company 1, its allocated dummy jobs cannot be more than $\phi^d_1$, simply because there are in total $\phi_K-\phi_1=\phi_1^d$ dummy layers in $G'$ and any company can only get at most one job from each dummy layer. Take an arbitrary company $i$, 
$2 \leq  i \leq K$, 
WLOG, suppose company $i$ is assigned $\phi^d_i +1$ dummy jobs. This extra one dummy job has to  come from another company $j$ ($j<i$) who has more dummy capacity than $i$. Hence, $j$, $j<i$, should receive $\phi^d_j - 1$ dummy jobs. If $j$ still gets $\phi^d_j$ dummy jobs, it blocks the possibility for $i$ to receive one extra dummy job as there will be not a sufficient number of dummy jobs in the dummy layers to support this allocation, due to the construction of dummy jobs in dummy layers.

Now, if $\phi^d_j,~j<i,$ will decrease, then there are two possibilities. First, if $0 \leq \phi^d_j-\phi^d_i \leq 1$, then $\boldsymbol{\phi}^d$ will not change as it will retain its order. Second, if $\phi^d_j-\phi^d_i \geq 2$, then this will result in a more even distribution of dummy jobs over companies. However, since $\boldsymbol{\phi}_f = \boldsymbol{\phi}' - \boldsymbol{\phi}^d$, this will result in a more even, or in other words, fairer distribution of jobs in $\boldsymbol{\phi}_f$. This contradicts our claim that $\boldsymbol{\phi}_f$ is max-min fair. The same reasoning holds if we decrease $\phi^d_i$, for $i= 1,\dots,K-1$ due to symmetry. If we decrease $\phi^d_i$, our only option is to increase a $\phi^d_j,~j<i$, which, as we have seen before, cannot occur due to insufficiently available dummy jobs.

\end{proof}

\begin{lemma}\label{lemma_cost}
The optimal solution for problem $P$ in terms of cost is the same as the optimal solution for problem $P'$.
\end{lemma}

\begin{proof}
Assume that the cost of the allocation of jobs in $P'$ is different than in $P$. If the allocation in $P'$ has a lower cost than the allocation in $P$, then because the dummy jobs have a cost of zero and we have added sufficient capacity, we can remove the dummy jobs and obtain an allocation for $P$ that has a lower cost than the optimal allocation in $P$. This contradicts the assumption that the allocation in $P$ is optimal. Now if the allocation in $P'$ has higher costs than the allocation in $P$, then we can add dummy jobs to the optimal allocation in $P$ and increase the capacities accordingly so that we obtain problem $P'$. We will then have an allocation for $P'$ that has a lower cost than the optimal allocation previously found in $P'$. This contradicts the assumption that the allocation in $P'$ is optimal. Therefore, the cost of the optimal allocation of jobs in $P$ is the same as the cost of the optimal allocation in $P'$.
\end{proof}

We now show the running time complexity of the proposed FairMinCost algorithm. After constructing the updated graph $G'$, we use a standard solution method for minimum cost maximum flow problems, namely the cycle-cancelling algorithm. Given a feasible flow, the cycle-cancelling method tries to find a negative cycle in the residual graph whose residual capacity it increases, so that the negative cycle disappears and the resulting solution is a solution with lower costs. Instead of choosing an arbitrary negative cycle, the cycle with the minimum mean cost is chosen, which makes the problem strongly polynomial-time solvable \citep{goldberg1989finding}. In order to find minimum mean-cost cycles, we use Karp's algorithm \citep{karp1978characterization}. This will give us our desired solution in which we have an allocation that is max-min fair and has minimum cost among all possible max-min fair allocations.

The runtime complexity of the FairMinCost algorithm is given below (see \ref{app:runtime2} for the proof).
\begin{theorem}\label{lemma_mfmca_runningtime}
The FairMinCost algorithm runs in time $O(J^3 K^3 (K+T)^3 (JK+JT+KT)^2 \log(JK+JT+KT))$.
\end{theorem}


As a conclusion, we can optimally solve MFMCA in polynomial time using first the IMaxFlow algorithm and then the FairMinCost algorithm.

\section{Computational results}\label{sec:computational}

In this section we investigate the performance of the algorithms through numerical experiments. We are interested in the following two performance measurements:

\begin{enumerate}
\item  The effect of fairness on the cost, the so-called \emph{price of fairness} (POF) \citep{bertsimas2011price}. POF is defined as the relative increase in the total cost (TC) under the fair solution, compared to the minimum cost (MC) solution; that is,
\[
POF = \frac{TC(\text{MFMCA}) - TC(\text{MC})}{TC(\text{MC})}.
\]

\item The effect of fairness on the job distribution. The job distribution depicts the number of jobs assigned to each company.
\end{enumerate}

Therefore, we generate test cases with various parameters and compute both the minimum cost solution (MC) using the standard minimum mean-cost cycle-cancelling algorithm, and the fair minimum-cost solution using the two-stage algorithm: first IMaxFlow, and then FairMinCost. Next to the costs, we take a look at the allocations in both cases and see how fairness influences the allocation.
All algorithms are coded in Java with the support of the JGraphT library \citep{jgrapht}. We run the experiments on the Lisa Compute Cluster of SURFsara \citep{lisasurfsara}.

\subsection{Test instances}
We derive test instances from our motivating example. In order to make the experiments representative of the situation at the Rotterdam port, we need representative values for the different parameters.
First of all,
we define one time unit as one hour, just as in the port. Although some tasks require more time than others, a task from one end of the port to the other does not take more than one hour due to the layout of the port.
We choose a time window of 10 time periods, $t_1, \ldots, t_{10}$, corresponding to a typical working day.
We then set the number of jobs to 250~\citep{duinkerken2006comparing}, and the number of companies to 50, based on the members of the ``VZV'' (Verenigde Zeecontainer Vervoerders), the Dutch alliance of sea-container carriers, which represents the different carriers in meetings with the terminals, the port, and other entities.

The jobs have a latest completion time. This is set to 3 time periods after the earliest time the job becomes available. This means that if a job becomes available at the beginning of $t_1$, it can be started at $t_2$ and $t_3$ as well but not at $t_4$, as its latest completion time was at the start of $t_4$.
Jobs are distributed over the 10 time periods but not uniformly. Since there are two peak hours throughout the day~\citep{duinkerken2006comparing}, we configure jobs to have a 25\% chance of starting at $t_2$ and another 25\% chance of starting at $t_6$. If a job would not specifically start at a peak hour, it has an equal chance to start at any time period from $t_1$ to $t_8$. This ensures that each job has a time window of 3 in which it can be executed. This also ensures that the number of jobs available in the first and last two time periods are smoothed out.

\paragraph{Scenarios.} In our experiments we assign each company a certain probability to bid on each job at an available time period. We distinguish three different scenarios. The first scenario is when all companies are not very eager to bid on jobs or they do not have many trucks available. In this \emph{low competition} scenario there is a 25\% chance of a company bidding on a job, for all companies, for all jobs. The second scenario is the exact opposite: companies are actually eager to bid on the jobs or they have many trucks available. In this \emph{high competition} scenario there is a 75\% chance of bidding on a job. In the third scenario we combine the first two scenarios by splitting up the companies into two groups of equal size, where the companies in the first group have low competition, and the companies in the second group have high competition. This case is more representative of reality, as there will be large companies that have plenty of trucks available, and there will also be smaller companies that only have a few trucks available. We call this the \emph{mixed competition} scenario.

The number of trucks all companies will have available at a certain time period, the \emph{capacity}, will be drawn uniformly random between 0\% and 5\% (\emph{5\% capacity}) or between 0\% and 10\% (\emph{10\% capacity}) of the total number of jobs they bid on in that particular time period, rounded to the nearest integer. Furthermore, we ensure that companies will have at least one truck available when they have bid on at least one job. Note that due to the dependency on the number of jobs they bid on in a time period, low competition companies will have fewer trucks available at a certain time period because they bid on fewer jobs, whereas high competition companies will have more trucks available because they bid on more jobs.

Now that we know how companies will bid on jobs, the question remains how much they will bid. We will have two cases here. The first case is when all companies have their bid drawn from the same distribution. We choose a uniform distribution that ranges from 30 to 60. We choose this range because the hourly wage of a truck driver plus the fuel costs for the largest distance within the port amounts to roughly 30 euros. Because companies would like to make some profit with these extra jobs, we let the bids range up to 60. We call this cost scenario the \emph{homogeneous costs} case and this can be applied to all three aforementioned bidding scenarios.
The second case is when some companies decide to bid relatively low while others decide to bid relatively high. In the cases of low and high competition, half of the companies will have their bids drawn from a uniform distribution that ranges from 30 to 50 and the other half from 40 to 60. When there is mixed competition, the low competition companies will bid high, from 40 to 60, whereas the high competition companies will bid low, from 30 to 50. The thought behind this is that low competition companies value their trucks more than the high competition companies do. Low competition companies only have a few trucks available and thus can only bid on a few jobs, whereas the high competition companies have more trucks available and will bid on more jobs. Therefore, the high competition companies will have to compete with many others for the same jobs, and so they will offer lower prices. We call this cost scenario the \emph{heterogeneous costs} case.

To summarize, we have six scenarios in total, each with 50 companies, 250 jobs, and a time window of 3 time periods for a job:
\begin{itemize}
\item[1.] Low competition, homogeneous costs (low/hom).
\item[2.] Low competition, heterogeneous costs (low/het).
\item[3.] High competition, homogeneous costs (high/hom).
\item[4.] High competition, heterogeneous costs (high/het).
\item[5.] Mixed competition, homogeneous costs (mix/hom).
\item[6.] Mixed competition, heterogeneous costs (mix/het).
\end{itemize}

Out of these six scenarios, we believe the last scenario with heterogeneous companies and heterogeneous costs to be the most interesting, because it comes closest to the real situation at the port. We also expect this scenario to yield a relatively bad performance in terms of price of fairness, because in the minimum cost solution most jobs will be allocated to the companies with high competition and low costs. However, because we want to enforce fairness, we need to also allocate jobs to the companies with low competition and high costs, which may increase the total cost substantially.

For each scenario, we run 100 experiments with both the minimum mean-cost cycle-cancelling algorithm for a minimum-cost solution and the fair two-stage algorithm for a fair solution. We record the job allocations in both solutions and the difference in total cost between the minimum-cost solution and the fair solution.

In the end we will investigate the effect of the amount of jobs and companies on the solutions. One can imagine that the price of fairness will differ depending on the number of jobs that needs to be allocated. To test this, we run experiments with 50 to 500 jobs in increments of 50, while maintaining all other parameters. In the same vein, the price of fairness may be dependent on the number of companies present. When there are only a few companies present the allocation may lack flexibility, whereas having many companies might drive up the costs because more companies have to be allocated a number of jobs. Therefore, we run experiments with 25 to 100 companies in increments of 25, while keeping the other parameters the same as in the base case.

\subsection{Results: price of fairness}
\begin{figure}
\centering
{\includegraphics[scale=0.6]{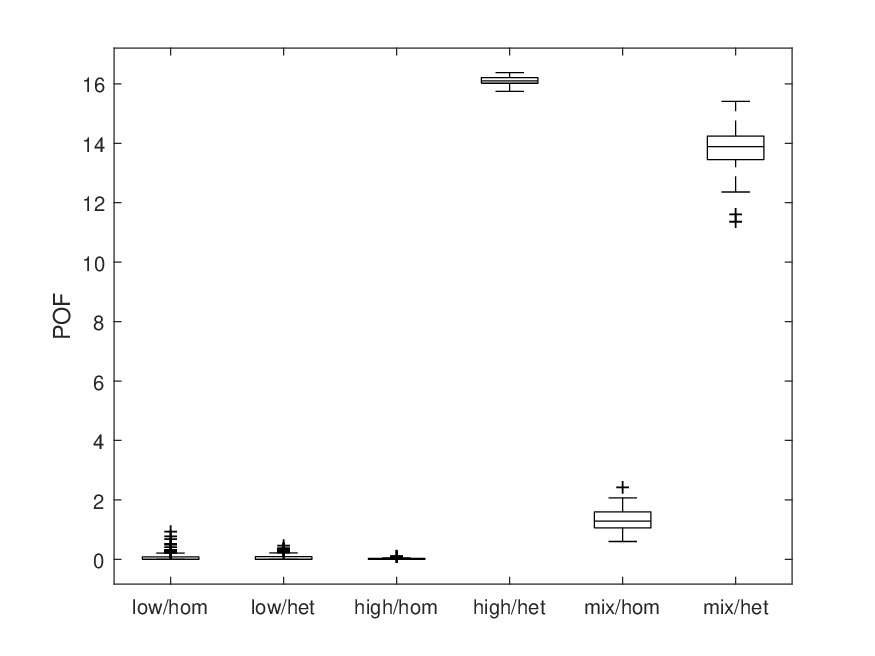}}
\caption{Boxplot of the price of fairness with 5\% capacity, with the different scenarios and the price of fairness on the horizontal and vertical axes, respectively.
\label{fig:gaps5}}
\end{figure}

\begin{figure}
\centering
{\includegraphics[scale=0.6]{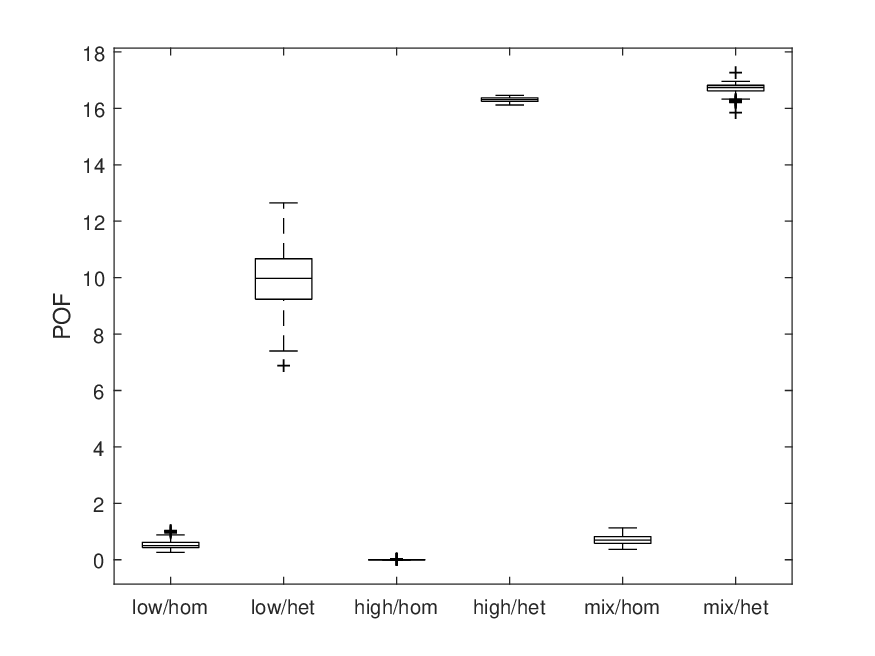}}
\caption{Boxplot of the price of fairness with 10\% capacity, with the different scenarios and the price of fairness on the horizontal and vertical axes, respectively.
\label{fig:gaps10}}
{}
\end{figure}

We first present the results of the fair allocations compared to the minimum-cost allocations for the experiments with all six scenarios.
In Figures \ref{fig:gaps5} and \ref{fig:gaps10}, the average price of fairness over 100 experiments per scenario, with \emph{5\% capacity} and \emph{10\% capacity}, respectively, are shown in a boxplot. Tables \ref{table:avgOptFair5} and \ref{table:avgOptFair10} show the accompanying statistics, i.e., the mean, standard deviation, and the minimum and maximum.

\begin{table}
\centering
{\scriptsize
{\begin{tabular}{|l|l|r|r|r|r|r|r|}
\hline
 			&			& low/hom 			& low/het 			& high/hom 					& high/het 			& mix/hom 			& mix/het \\
\hline
Min-cost 	& mean	& 6477.97  	& 7362.67  	& 7515.21  & 7539.31  	& 7559.42  	& 7537.85 	\\
			& std & 355.28 & 424.07 & 4.06 & 8.41 & 8.20 & 8.11 \\
			& min & 5585 & 6236 & 7506 & 7522 & 7535 & 7250	\\
			& max & 7335 & 8417 & 7526 & 7560 & 7579 & 7564	\\
\hline
Fair  		& mean	& 6483.29 	& 7367.40 	& 7516.45 & 8752.90 & 7658.93 & 8578.70	\\
			& std & 358.25 & 425.22 & 4.52 & 4.12 & 28.23 & 51.025 \\
 			& min & 5582 & 6245 & 7506 & 8732 & 7602 & 8403	\\
 			& max & 7385 & 8417 & 7529 & 8760 & 7743 & 8695	\\
\hline
POF			& mean	& 0.08 & 0.06 & 0.02 & 16.10 & 1.32 & 13.81 \\
			& std & 0.17 & 0.11 & 0.02 & 0.13 & 0.37 & 0.70 \\
			& min & 0.00 & 0.00 & 0.00 & 15.75 & 0.60 & 11.36 \\
			& max & 0.93 & 0.46 & 0.11 & 16.38 & 2.42 & 15.41 \\
\hline
\end{tabular}}
}
\caption{Statistics of minimum cost and fair cost and POF over 100 experiments with 5\% capacity.
\label{table:avgOptFair5}}
{}
\end{table}

\begin{table}
\centering
{\scriptsize
{\begin{tabular}{|l|l|r|r|r|r|r|r|}
\hline
 			&			& low/hom 			& low/het 			& high/hom 					& high/het 			& mix/hom 			& mix/het \\
\hline
Min-cost 	& mean	& 7708.83	& 8069.41 & 7509.81 & 7524.81 & 7535.81 & 7524.44	\\
			& std & 22.24 & 90.82 & 3.45 & 5.26 & 6.70 & 5.80 \\
			& min & 7661 & 7857 & 7503 & 7514 & 7518 & 7511	\\
			& max & 7774 & 8294 & 7522 & 7538 & 7557 & 7542	\\
\hline
Fair  		& mean & 7750.36 & 8872.97 & 7509.90 & 8751.89 & 7589.32 & 8780.72	\\
			& std & 25.92 & 17.68 & 3.48 & 1.39 & 14.12 & 12.84 \\
 			& min & 7704 & 8833 & 7503 & 8750 & 7561 & 8733	\\
 			& max & 7816 & 8934 & 7522 & 8756 & 7622 & 8815	\\
\hline
POF			& mean	& 0.54 & 9.97 & 0.00 & 16.31 & 0.71  & 16.70 \\
			& std & 0.16 & 1.19 & 0.00 & 0.08 & 0.17 & 0.20 \\
			& min & 0.26 & 6.88 & 0.00 & 16.12 & 0.37 & 15.85 \\
			& max & 1.03 & 12.65 & 0.03 & 16.46 & 1.13 & 17.27 \\
\hline
\end{tabular}}
}
\caption{Statistics of minimum cost and fair cost and POF over 100 experiments with 10\% capacity.
\label{table:avgOptFair10}}
{}
\end{table}

It is clear that costs play an important role in the differences in the total cost between the minimum-cost and fair solutions. In the scenario with homogeneous costs (i.e., low/hom, high/hom, mix/hom), the price of fairness ranges from $0$ to $2.42$. However, in the scenarios with heterogeneous costs (i.e., low/het, high/het, mix/het) the price of fairness ranges from $0$ to a staggering $17.27$.
This is as expected, because if the costs are similar for all jobs for all companies, it will be relatively easy to reallocate jobs to a different company with similar costs. Once costs vary more among companies there will be an increase in costs because jobs that were allocated to relatively cheap companies are forced to be reallocated to more expensive companies.

The seeming discrepancy in the low/het scenario between the $5\%$ and $10\%$ capacity cases can be explained by the structure of the low/het scenario. Because there are only low competition companies in this scenario, all companies will bid on only a few jobs. When we set the capacity for each time period to only $5\%$ of those bids, it happens frequently that the capacity is set to an extremely low number, i.e.,  $0$ or $1$. This means that there is not much leeway in the fair solution for jobs to be reallocated. The fair solution is often similar to the minimum-cost solution as there are only a few other possible allocations. This can also be observed in Table \ref{table:avgOptFair5} where the means of the costs of the minimum-cost and the fair solutions in the low/het scenario are similar but are extremely high compared to those in the low/hom scenario.

By increasing the capacity to $10\%$, we allow more room for jobs to be reallocated. In Table \ref{table:avgOptFair10} we can see that the difference in average cost between the low/het and low/hom scenarios in the minimum-cost solution is significantly smaller than in the $5\%$ capacity case. As there is more room for reallocation, this eventually results in a higher price of fairness.

We can see that competition also has an influence on the price of fairness. Both high and mixed competitions result in a higher price of fairness than low competition. We can see this in the low/het, high/het, and mix/het scenarios in the $10\%$ capacity case (average price of fairness: $9.97$, $16.31$, and $16.70$ respectively). At first this may seem surprising. One would imagine that having more possibilities for allocation will result in both the minimum-cost and fair solutions to be closer to each other compared to when there are limited possibilities. However, this discrepancy can be explained by looking at the minimum cost and fair cost of the solutions (see Table \ref{table:avgOptFair10}). We can see that the cost for the minimum-cost solutions in the low/het scenario is on average higher ($8069.41$) than that of the high/het ($7524.81$) and mix/het ($7524.44$) cases. This is due to the limited possibilities if there are companies bidding only on a few jobs. However, the average cost in the fair solutions in the low/het scenario ($8872.97$) is similar to that of the high/het ($8751.89$) and mix/het ($8780.72$) scenarios. This results in a smaller difference between the minimum-cost and fair solutions in the low/het scenario compared to the high/het and mix/het scenarios.

The mix/het scenario, where there is a mix of low- and high-competition companies, seems to have a price of fairness similar to or lower than the high/het scenario ($13.81$ against $16.10$ in the $5\%$ capacity case, and $16.70$ against $16.31$ in the $10\%$ capacity case). This is somewhat surprising at first. Due to the presence of high-competition companies, it is clear that the minimum-cost solutions would be similar to the solutions in the scenario with high competition because the more expensive low-competition companies are being ignored.
However, one would expect that the fair solutions will have higher costs because the more expensive low-competition companies also need to be allocated jobs. For this we have to keep in mind that the degree of fairness is not equal among the scenarios. It appears that due to the presence of low-competition companies, which have fewer bids and therefore fewer allocation possibilities, they get fewer jobs allocated to them even in the fair allocation. This in turn means that the other, high-competition companies get allocated more jobs that are cheaper. In the end, this results in lower costs overall. This effect can be seen clearly in the $5\%$ capacity case.
When we increase the possible capacity to $10\%$, low-competition companies get assigned more jobs, almost as many as the high-competition companies. This results in higher costs.

\paragraph{\textbf{Summary}.}
All things considered, we can see that the price of fairness is fairly low when the costs are homogeneous among companies. Jobs can be easily reallocated to make a more fair allocation while keeping the total cost similar because the individual costs of a job for each company are similar. In this case, fairness can be easily enforced without increasing the costs too much or at all. When costs are heterogeneous however, we have to pay a higher price for fairness. This is as expected because we would also need to allocate jobs to companies with high costs, whereas we would only opt for companies with low costs in the minimum-cost solution.
In the case of low competition, allocations tend to have slightly higher costs because there is a limited availability of jobs to be allocated to companies. This holds true for both the minimum-cost and fair allocations.
In the cases of high and mixed competitions, the costs of the minimum-cost solutions are similar in the cases with homogeneous costs. This is as expected because only the companies with the lowest costs get chosen while the ones with high costs are ignored. However, when we enforce fairness, jobs will be forcibly reallocated to companies with high costs, which might increase the total cost.
The price of fairness is the highest in the high/het and mix/het scenarios. Depending on the actual discrepancy between the different costs, one might opt out of the idea of enforcing fairness when the price of fairness becomes too high.

\subsection{Results: job distribution}
\begin{figure}
\centering
{\includegraphics[scale=0.6]{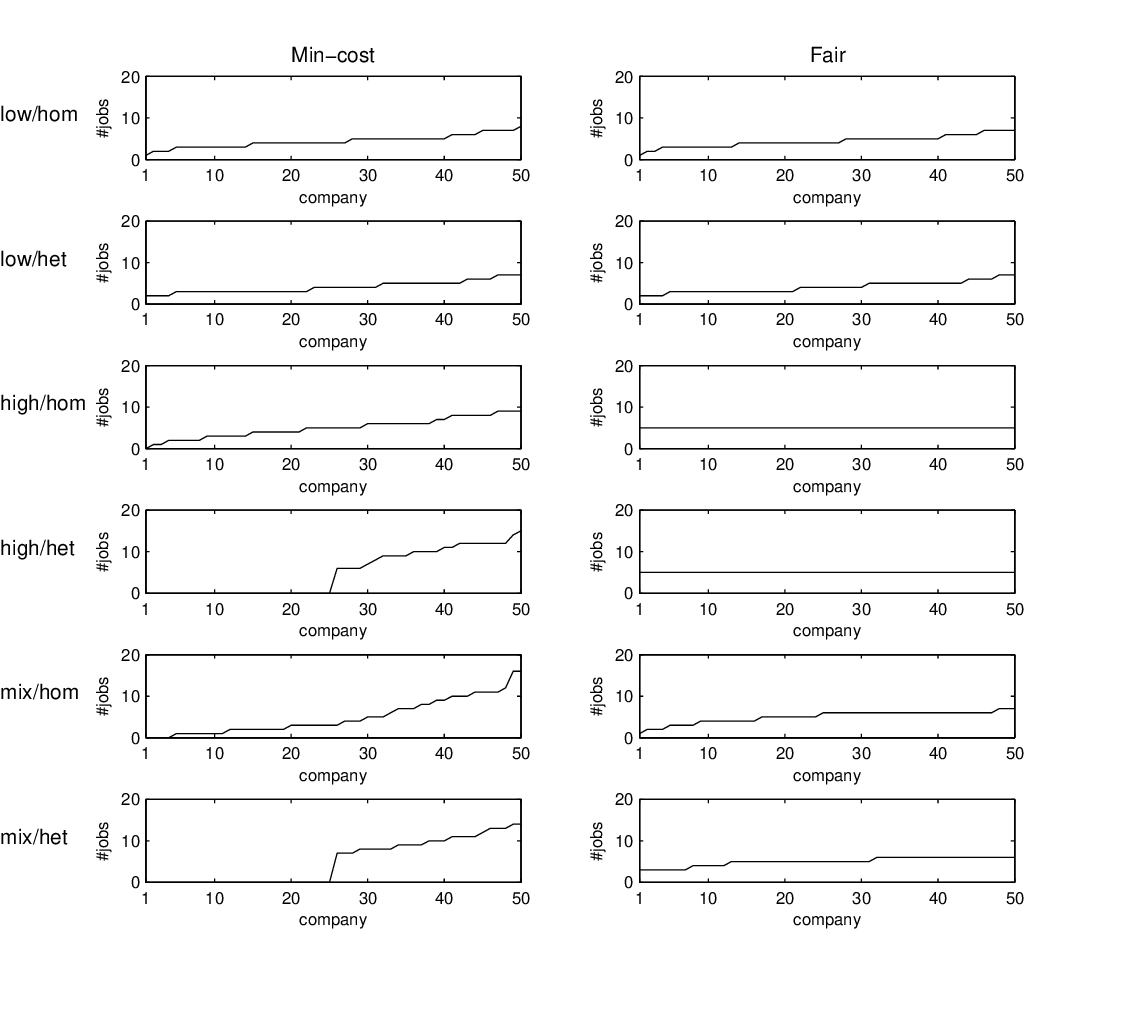}}
\caption{Job distributions in the minimum-cost and fair solutions for each scenario with 5\% capacity, sorted in ascending order by number of assigned jobs, with companies on the horizontal axis and the number of assigned jobs on the vertical axis.\label{fig:distributions5}}
{}
\end{figure}

\begin{figure}
\centering
{\includegraphics[scale=0.6]{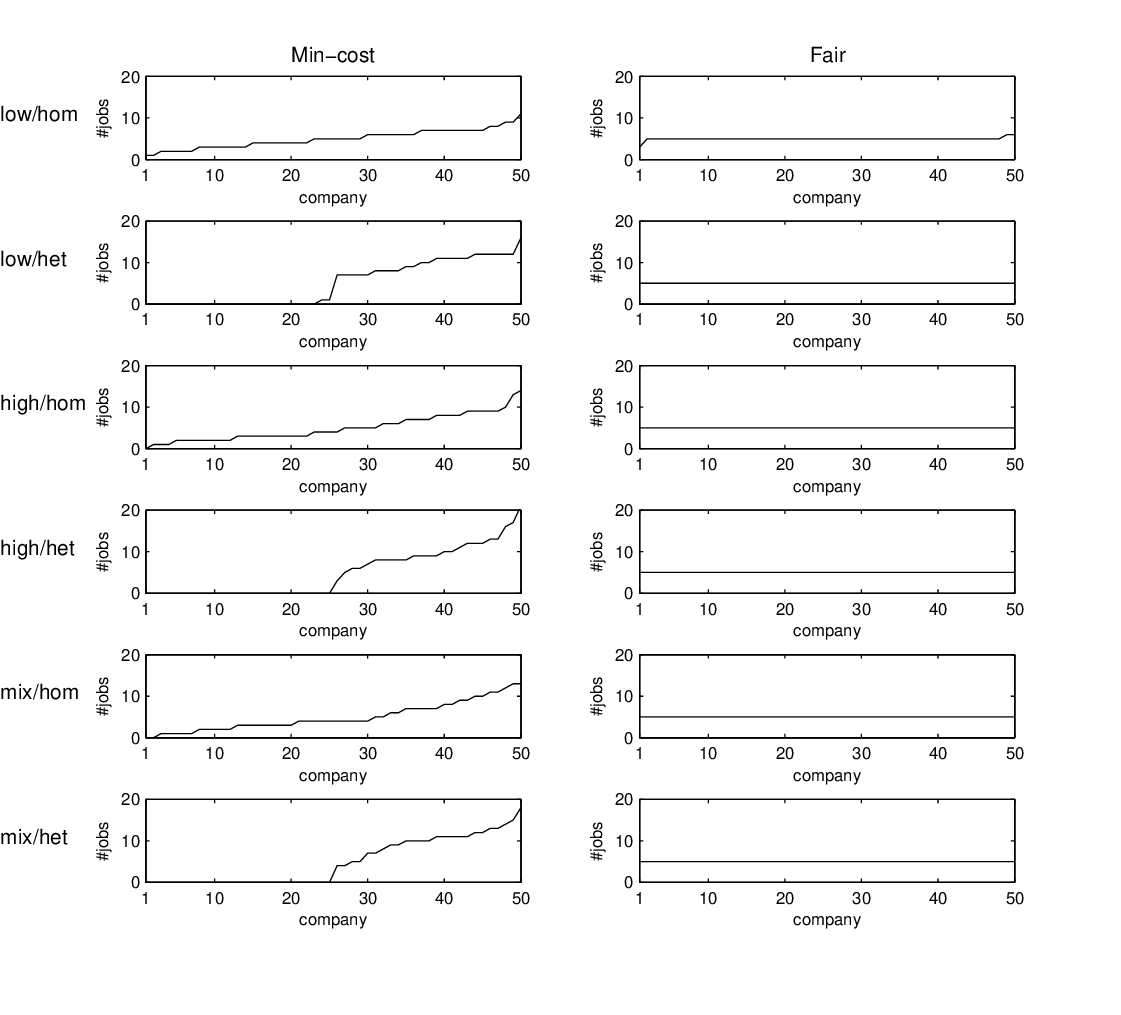}}
\caption{Job distributions in the minimum-cost and fair solutions for each scenario with 10\% capacity, sorted in ascending order by number of assigned jobs, with companies on the horizontal axis and the number of assigned jobs on the vertical axis.
\label{fig:distributions10}}
{}
\end{figure}

Figures \ref{fig:distributions5} and \ref{fig:distributions10} show job distributions of both the minimum-cost and the fair solutions at one instance for each scenario. We choose to show the job allocations of the instances in which the highest POF was found because there were many cases with a POF of $0\%$ in some scenarios. The allocations are sorted in nondecreasing order so that they represent the fairness vector. We can see that in all cases the fair job distribution is smoother than the minimum-cost distribution, which is exactly what we desire.

For the low/hom and low/het scenarios in the $5\%$ capacity case, we can see that the job distributions in the minimum-cost and fair solutions do not differ much, with the exception of a few companies getting one job more, or less. This further supports our claim that in these scenarios it is often the case that the fair solution is similar to the minimum-cost solution because there exists only a few feasible allocations.

In the high-competition scenarios (high/hom and high/het) for both the $5\%$ and $10\%$ capacity cases, we can see that the fair allocation assigns each company the same number of jobs. This is possible because all companies have many bids, which results in much leeway while reallocating.
In the low- and mixed-competition scenarios we can see that due to the inclusion of low-competition companies that do not bid on many jobs, it can still happen that certain companies get assigned fewer jobs than others. This is due to capacity restrictions and to the fact that it is simply not feasible for our proposed algorithm to assign more jobs to those companies. This effect can therefore be seen to be more prominent when there is a lower capacity ($5\%$).

In the homogeneous cost scenarios (low/hom, high/hom and mix/hom) of the $10\%$ capacity case, we can see that in the minimum-cost solution there are a few companies that do not get assigned any jobs or are assigned only a few jobs. This is primarily due to their higher individual costs compared to other companies. However, in the fair solution all companies are allocated roughly the same number of jobs. A few companies will receive fewer jobs than others, but this is due to capacity restrictions as explained above. Given that there are no or just minimal differences between the costs of the minimum-cost and the fair solutions in the homogeneous cost scenarios, the price of fairness when costs are homogeneous is minimal. This is as expected. Reallocating jobs does not come at a significant price because the cost for a job is similar among all companies.
A similar effect can be seen in the homogeneous cost cases in the $5\%$ capacity case. However, not all companies get the same number of jobs due to capacity restrictions being more prominent.

With heterogeneous costs, the first 25 companies in the minimum-cost solution have been allocated only a few jobs or even none at all because of the higher costs these companies have (except for the $5\%$ capacity low/het scenario, which is due to capacity restrictions). However, in the fair solution, these companies do get a significant number of jobs, although this comes with a hefty POF, which can get up to as much as $17.27$.

\subsection{Results: varying number of jobs}
\begin{figure}
\centering
{\includegraphics[scale=0.6]{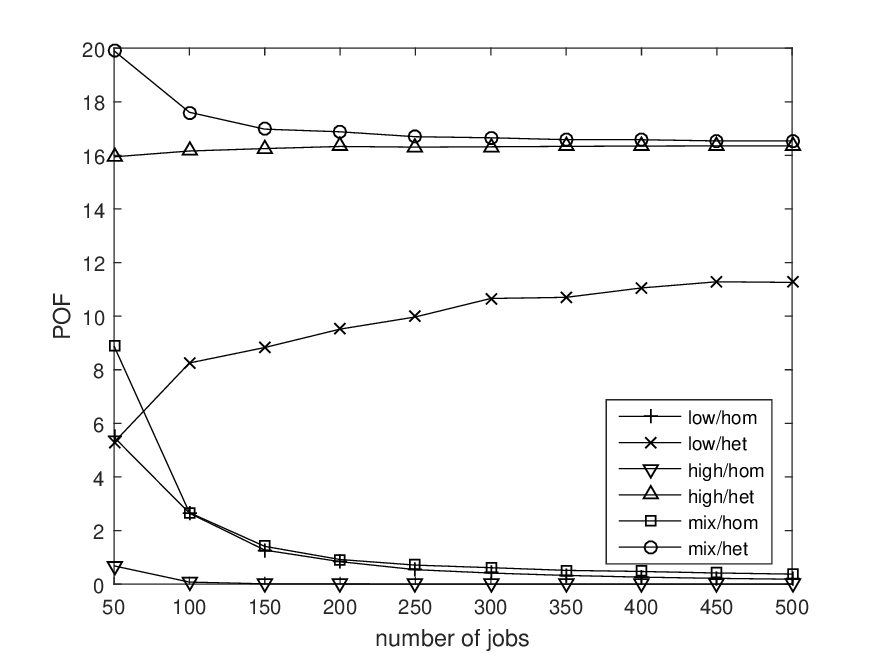}}
\caption{Line charts of average POF over various numbers of jobs with 10\% capacity, with the number of available jobs in an experiment and the average price of fairness over 100 experiments on the horizontal and vertical axes, respectively.
\label{fig:varyingjobs}}
{}
\end{figure}

We investigate the effect of the number of jobs on the price of fairness. We vary the number of jobs from 50 to 500 while keeping the same number of companies of 50. For each scenario and number of jobs we run 100 experiments and take the average price of fairness over these experiments. The results are displayed in Figure \ref{fig:varyingjobs}.

We can see that for scenarios with homogeneous costs the price of fairness can be rather high when the number of jobs is low, as much as $8.88$ in the mix/hom scenario. This can be accredited to the lack of flexibility in allocation when there is a small number of jobs. The distribution from which the costs of jobs are drawn may be the same for all jobs, but there is still variation in the costs. With a small number of jobs this variation plays a larger role when reassigning jobs from the minimum-cost solution to a fair solution, especially when low-competition companies that bid on only a few jobs are present. Even if low-competition companies had a high bid on a particular job, if it is only one of the few jobs they bid on, the job needs to be allocated to these same bidders in the fair solution. This raises the total cost in the fair solution.
We notice this effect particularly in the mix/hom scenario. A job that was assigned to a high-competition company in the minimum-cost solution, where the lowest cost was chosen, can suddenly be assigned to a low-competition company that only submitted a few bids.

As the number of jobs increases, the number of bids also increases, providing more flexibility for reallocation when a fair solution needs to be constructed. There will be a bigger chance of having a bid for the same job from another company with similar cost as the company in the minimum-cost solution. The average price of fairness decreases when the number of jobs increases and eventually the price of fairness seems to stabilize close to zero.

For scenarios with heterogeneous costs the effects are more complicated. Due to heterogeneous costs, jobs that were allocated to cheap companies (ones that bid between $30$ and $50$) in the minimum-cost solution need to be reallocated to expensive companies (ones that bid between $40$ and $60$) in the fair solution. This means an average increase of $10$ in the cost per reallocated job. When the number of jobs increases, the share of reallocated jobs increases as well, thus increasing the total cost.
This effect can be seen in particular in the low/het scenario, where the average price of fairness increases as the number of jobs increases. In the high/het scenario this effect is not as prominent, because there is high competition and thus there are many bids to choose from. Then there is a substantial chance that there exists a bid from another company that is similar in cost.

For the mix/het scenario, we have to keep in mind that high-competition companies have bids between $30$ and $50$, whereas low-competition companies have bids between $40$ and $60$. This means that when a job from a high-competition company in the minimum-cost solution needs to be reassigned to a low-competition company in the fair solution (something that also happens in the mix/hom scenario) the cost will increase by $10$ on average. This results in a less steep decline in the average price of fairness compared to the mix/hom scenario when gradually going from $50$ jobs to $200$ jobs. The increase in costs resulting from reallocating jobs to companies with completely different costs is especially noticeable when there are fewer jobs. This increase in costs due to reallocation weighs more in the mix/het scenario than the effect of the average increase of $10$ when switching from a high-competition company to a low-competition one. The average price of fairness is therefore declining in the mix/het scenario in contrast to the low/het and high/het scenarios where the price of fairness increases with the increase in the number of jobs.

For all scenarios with heterogeneous costs it seems that the average price of fairness eventually stabilizes as the number of jobs increases. This again can be accredited to the flexibility that the increasing number of jobs, and therefore bids, introduces for reallocation when a fair solution needs to be constructed. Reallocations become more efficient and will eventually weigh up against the increase in costs due to the increased amount of reallocations.


\subsection{Results: varying number of companies}
\begin{figure}
\centering
{\includegraphics[scale=0.6]{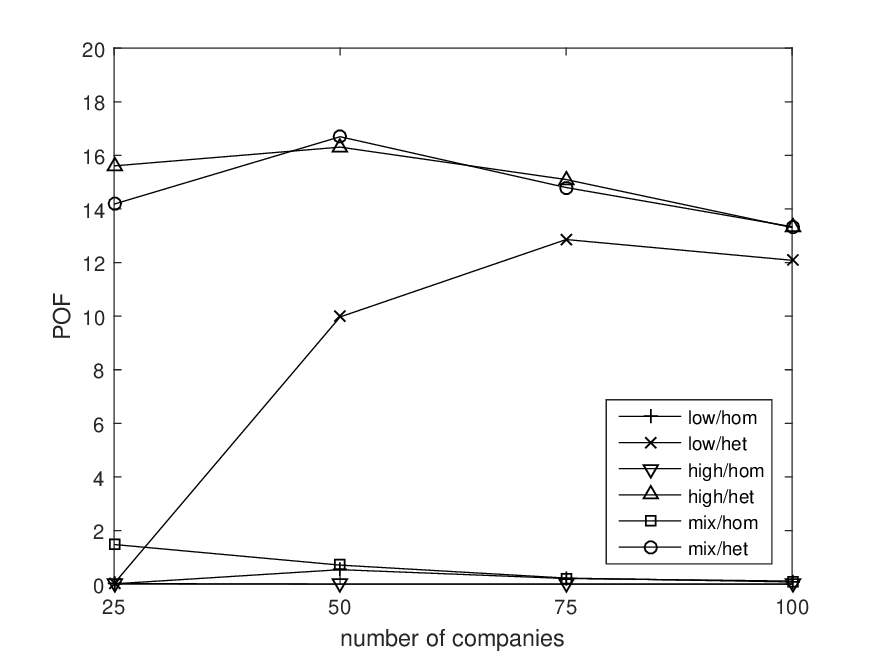}}
\caption{Line charts of average POF over various numbers of companies with 10\% capacity, with the number of companies in an experiment and the average price of fairness over 100 experiments on the horizontal and vertical axes, respectively.
\label{fig:varyingcompanies}}
{}
\end{figure}

In order to investigate the effect of the number of companies on the price of fairness we vary the number of companies from 25 to 100 in increments of 25, while now fixing the number of jobs to 250. For each scenario and number of companies we run 100 experiments and again take the average price of fairness. The results are shown in Figure \ref{fig:varyingcompanies}.

We notice that for the scenarios with homogeneous costs the price of fairness is rather low, ranging from 0.00 to 1.48 in the mix/hom scenario. The addition of more companies does not have much effect, as the costs are similar among all companies. This creates more flexibility for the fair allocation. Only when the number of companies is low in the mix/hom scenario, there is a slightly higher price of fairness due to lack of flexibility to allocate the jobs to companies with similar costs in the fair allocation. 

When we look at the scenarios with heterogeneous costs, we can see that for the high/het and mix/het scenarios the price of fairness seems to decrease as more companies participate, which is again due to added flexibility as the same number of jobs can be distributed among more companies with many more bids. The seemingly odd occurrence of a slightly lower price of fairness at 25 companies compared to the case with 50 companies can be explained by the lack of competition. When we take a look at the actual costs of the minimum cost allocations, we can see that it is slightly higher in the case of 25 companies than it is when there are 50 companies. Because of the limited number of companies, there is not much competition between the bids. However, the costs for the fair allocations with 25 companies is similar to those in the cases with more companies, yielding a lower price of fairness.

The results of the low/het scenario stand out the most. It seems that the minimum cost and the fair allocations have similar costs when there are 25 companies, having an average price of fairness of 0.02. The price of fairness then increases substantially to 9.97 and 12.86 as the number of companies increases to 50 and 75 companies, respectively. It decreases again to 12.08 when the number of companies is further increased to 100.
The average price of fairness of 0.02 with 25 companies can be easily explained when we take a look at the allocations. It seems that due to the small number of bids with only 25 companies and low competition, the fair allocation is often exactly the same as the minimum cost allocation. There is simply no other allocation possible.
As the number of companies increases, the number of bids also increases, adding more leeway for the fair allocation. With 50 companies the number of bids seems to be sufficient in order to distribute the jobs to companies evenly. However, the number of bids is still relatively low, resulting in the costs of the minimum cost allocations to be much higher than in the case of 75 or 100 companies. At the same time, the costs of the fair allocations gradually decrease as the number of companies increases. The decrease in costs of the minimum cost allocations is much heftier than the decrease in costs of the fair allocations, which results in the increase in price of fairness. Going from 75 to 100 companies the number of bids increases again, lowering the costs for the minimum cost allocations slightly, while the costs for the fair allocations decrease more with the added bids. This finally results in a slight drop in the price of fairness.

\section{Conclusions and discussion}\label{sec:con}
In task or job allocation problems there are many ways to assign jobs to all interested parties. The most common way is to minimize the costs of such allocation by only considering the cheapest companies.
In this paper, instead of just focusing on costs, we take into account the job distribution over companies. We try to allocate jobs to all participating parties as fairly as possible in terms of the number of allocated jobs. This additional criterion is particularly relevant in our motivating example, an inter-terminal transport problem (ITT) in the port of Rotterdam, where we want to use the trucks already present at the port to execute inter-terminal transport jobs. Because such a job allocation will be repeated daily, it is crucial to give those companies incentives to be involved in this activity by assigning them jobs based not only on their costs but also on ensuring some market share, i.e., allocated jobs.

To meet the new fairness criteria in task allocation, we developed a polynomial-time optimal method consisting of two novel algorithms: IMaxFlow, which uses a progressive filling idea, and FairMinCost, which smartly alters the structure of the problem. The output of these two algorithms is a max-min fair task allocation with the least total cost. In the experiments we looked at several scenarios for both the jobs that are being auctioned, and the companies who are bidding on the jobs. From the results of the experiments we find that in the situation where the costs among companies are similar, implementing fair allocations comes with almost no extra cost for the task owner.
When the prices are highly volatile however, the auctioneer may need to pay more for the fairness.
When there are relatively few jobs, the price of fairness will usually be relatively high due to the lack of flexibility in reallocating jobs. As the number of jobs increases, the price of fairness will stabilize due to the flexibility granted by the increase in the number of bids.
Similarly, the more companies are participating, the more bids there will be, resulting in more flexibility for reallocation and a lower price of fairness. However, the price of fairness can also be low when there are only few companies. This is then mainly due to lack of flexibility for reallocation, so that the fair allocation is similar to the minimum cost allocation. This means that the number of participants should be sufficient in order to have the desired flexibility needed for reallocation.
The auctioneer should contemplate whether the fairness in the allocation is worth the extra cost.
It is necessary to investigate specific cases regarding the price of fairness.

We made certain assumptions in this work because we had a real case of the ITT problem in the port of Rotterdam in mind.
Some of these assumptions can be relaxed to some extent. For instance, we assumed that each task can be completed within one time unit. If the tasks have different durations, we can normalize them by using a time unit large enough to encompass the task with the longest duration. We may lose some efficiency by doing this, but it makes the problem solvable using our proposed algorithms.
Furthermore, we assume that the tasks are independent.
One way to tackle interdependent tasks is to make them available in subsequent time periods, i.e., make sure one task has been executed before the next one is made available for execution.

The focus of this research is on designing efficient algorithms for finding fairest task allocations. Auctions are used in this research as a way to collect local information from the participants. This information is then used as input in the task allocation problem. We do not consider the bidding behaviour of the bidders in this paper. However, bidders may
choose to misreport their inputs in an attempt to affect the allocation in their favor.
In order to incentivize the bidders to bid truthfully, the mechanism design aspect of the auction needs to be studied as future research~\citep{krogt08ecai}.
In addition, we have only looked at one quantification of fairness in this research, in which we only consider the number of tasks in an allocation. As we have seen in the literature, there are many definitions of fairness and many different quantifications of fairness~\citep{Ogryczak2008451,GonzalezPachon201647}. 
When considering game theoretical properties of the auction mechanism, another quantification of fairness might prove to be better in attaining the desired properties. Investigating different quantifications of fairness with mechanism design will be an interesting future direction.


In many real-world cases, the bids from one company can be combinatorial, that is, the cost of receiving two jobs is strictly smaller than the total cost of executing the same two jobs separately.
If this combinatorial property exists between jobs, the task allocation problem becomes NP-hard~\citep{cramton2006combinatorial}.
For example in the ITT problem, if one job is to transport some goods from location A to B, and  another job is to ship some goods from B to C, it seems that giving both jobs to one company leads to smaller costs than letting two companies execute the two jobs separately. After consulting with a port manager, we find out that the margin of these two instances is so narrow that we gladly ignored the combinatorial property. The advantage of this is that we now have a polynomial-time algorithm to compute the optimal allocation in terms of fairness and cost.
For the cases where the tasks have a high degree of complementarities our proposed algorithms cannot be directly applied. Adapting our algorithms for solving such cases is open for further research.

Even though our work has been inspired by the situation in the port of Rotterdam, the described setting is not unique to this application. This work can be applied to many other task allocation problems
in which the centralized planner wants to enforce some kind of fairness among the agents.
Due to the rise of the so-called sharing economy \citep{goldman2006sustainable, belk2014you}, collaborative consumption in transport, for example car-sharing, has gained interest in the past years. The main concern in this area is on where to station the shared-use vehicles \citep{fan2008carsharing, shaheen2010carsharing, kek2009decision}.
However, online platforms for collaborative consumption in transport have recently been upcoming. In these platforms participants are free to join or leave as they please.
One might think of taxi services that are operated by civilians.
Another application would be in airport slot allocation. In this area, although not as dynamic as in the cases of the port and taxi services, it is important to allocate slots to airlines both efficiently and fair, as such to motivate new entrants \citep{castelli2011airport, condorelli2007efficient}.
It would be interesting to see whether our methods can be used in those applications.

\appendix

\section{Proof of Theorem \ref{lemma_mlmf_runningtime}}
\label{app:runtime1}
\begin{proof}
IMaxFlow starts with capacity 0 for all company-sink edges and it adds only 1 more capacity at each iteration. Thus, IMaxFlow takes at most $\max_{k \in \mathcal{K}}(N_{k}) < J$ iterations, and in each iteration there are at most $K$ steps. In each step, a maximum flow algorithm is called. In our case, this is the Edmonds-Karp algorithm, which runs in time $O(\card{V}\card{A}^2)$. The flow network $G$ consists of at most 2 (source and sink) $+ J + JT + TK + K$ nodes and at most $J + JT + JTK + TK + K$ arcs. This results in a running time of $O(JK (JT+TK)(JKT)^2) = O((J^4 K^3 T^3) + (J^3 K^4 T^3))$.
\end{proof}

\section{Proof of Theorem \ref{lemma_mfmca_runningtime}}
\label{app:runtime2}
\begin{proof}
The Goldberg-Tarjan algorithm is known to terminate after $O(\card{V}\card{A}^2 \log(\card{V}))$ iterations, with Karp's algorithm having a running time of $O(\card{V}\card{A})$. This results in a $O(\card{V}^2\card{A}^3\log(\card{V}))$ algorithm for solving the second stage of the MFMCA problem.
In $G'$ there are $\phi_{K}-\phi_1 < J$ dummy layers. In each dummy layer the number of dummy jobs is upper bounded by $K$. The number of vertices in each dummy layer is then at most $3 K$. This results in the number of vertices being upper bounded by $(JT + TK)$ for the original problem $P$ and by $J K$ for the dummy part of problem $P'$, for a total of $JT + TK + JK$.
The number of arcs $\card{A}$ is upper bounded by $JTK$ for $P$ and by $J K^2$ for the dummy part of problem $P'$, for a total of $JTK + J K^2$.
Hence, FairMinCost runs in time $O(J^3 K^3 (K+T)^3 (JK+JT+KT)^2 \log(JK+JT+KT))$.
\end{proof}


\end{document}